\documentclass{article} 
\usepackage{iclr2027_conference,times}

\usepackage{amsmath,amsfonts,bm}

\def\eqref#1{equation~\ref{#1}}

\def\1{\bm{1}}

\DeclareMathAlphabet{\mathsfit}{\encodingdefault}{\sfdefault}{m}{sl}
\SetMathAlphabet{\mathsfit}{bold}{\encodingdefault}{\sfdefault}{bx}{n}

\usepackage{hyperref}
\usepackage{url}
\usepackage{algorithm}
\usepackage{algorithmic}

\usepackage{amsmath}
\usepackage{amssymb}
\usepackage{mathtools}
\usepackage{amsthm}
\usepackage{comment}
\usepackage{makecell}
\usepackage{booktabs}

\newcommand{\bbb}{\mathbf{b}}

\newcommand{\Ab}{\mathbf{A}}
\newcommand{\EE}{\mathbb{E}}
\newcommand{\RR}{\mathbb{R}}
\newcommand{\norm}[1]{\|#1\|}

\newcommand{\btheta}{\bm{\theta}}

\title{A Concentration Bound for Two-Timescale Actor-Critic Algorithm}

\author{Prashansa Panda  \\
Department of Computer Science and Automation\\
IISc Bangalore \\
\texttt{prashansap@iisc.ac.in} \\
\And
Shalabh Bhatnagar \\
Department of Computer Science and Automation \\
IISc Bangalore \\
\texttt{shalabh@iisc.ac.in} \\
}

\begin{document}

\ifx\theorem\undefined
\newtheorem{theorem}{Theorem}

\newtheorem{lemma}{Lemma}
\newtheorem{sublemma}{lemma}[lemma]

\ifx\remark\undefined
\newtheorem{remark}{Remark}

\ifx\proposition\undefined
\newtheorem{proposition}{Proposition}

\ifx\assumption\undefined
\newtheorem{assumption}{Assumption}

\ifx\corollary\undefined
\newtheorem{corollary}{Corollary}

\maketitle

\begin{abstract}
Significant research effort has been directed in recent years towards establishing both asymptotic and non-asymptotic convergence guarantees for two-timescale actor–critic algorithms, where the actor recursion is run on a slower timescale than the critic recursion. This work derives a uniform all-time concentration bound for the actor-critic  algorithm with function approximation in the long-run average-reward setting.  This bound helps us analyze the behavior of the actor parameter with high probability. We show that, after some finite time, the actor parameter enters a safe region and remains within it thereafter with a high probability. Specifically, with probability at least $1-\epsilon_1 - \epsilon_2$, the actor error $\Vert \theta_k - \theta^{*}\Vert$ is $O \bigg( \frac{n_0^{3/4}}{k} \frac{1}{\sqrt{\epsilon_2}}+   \bigg(\frac{1}{n_0}\bigg)^{1/4} \log^{1/4} \bigg( \frac{1}{\epsilon_1  }\bigg)   + \bigg(\frac{1}{n_0} \bigg)^{1/4} \bigg)$ for all $k \geq n_0 $ and sufficiently large $n_0$. We also present experimental results demonstrating that the aforementioned actor error diminishes with the number of actor-parameter updates. 
\end{abstract}

\section{Introduction}

Actor-Critic (AC) methods have demonstrated strong performance across a wide range of reinforcement learning (RL) problems. Pure policy-based approaches, such as REINFORCE, often encounter high variance when estimating policy gradients, while value-based methods like Q-learning perform well in tabular environments but may become unstable or diverge when combined with function approximation. AC methods address these limitations by integrating policy-based and value-based learning. In this framework, the actor aims to learn an optimal policy using feedback from the critic, whereas the critic estimates the value function corresponding to the actor’s current policy. The stability of these algorithms is typically achieved through a separation of learning timescales between the actor and the critic.

AC algorithms are based on coupled stochastic update rules that operate on two distinct timescales, with the actor’s learning rate generally decreasing more rapidly than the critic’s. This separation of timescales plays a fundamental role in maintaining the stability of the updates and guaranteeing their almost sure convergence. From the perspective of the faster timescale, the slower process appears nearly constant, whereas from the perspective of the slower timescale, the faster process can be viewed as having already converged. Such a mechanism allows AC algorithms to effectively mimic the behavior of the policy iteration procedure for numerically solving Markov decision processes (MDPs), thereby facilitating convergence toward an optimal policy. The asymptotic convergence properties of two-timescale AC methods are commonly analyzed using the ordinary differential equation (ODE) framework. 
In \citep{bhatnagar2023actorcritic}, a critic-actor (CA) algorithm was introduced for the lookup-table setting under the infinite-horizon discounted-cost criterion. Unlike conventional actor-critic methods, the algorithm reverses the roles of the actor and critic by interchanging their learning timescales. As a result, the resulting update scheme follows the dynamics of value iteration rather than policy iteration. 
Finite-time analysis and concentration bounds have recently gained prominence as approaches for establishing non-asymptotic convergence guarantees.
 Recent studies have also investigated the finite-time convergence properties of various actor–critic algorithms. Non-asymptotic (such as finite-time) analyses provide sample complexity guarantees and optimal learning rates, while concentration-bound analyses enable the derivation of high-probability convergence bounds.

In this paper, we derive a finite-time, high-probability convergence bound for the two-timescale actor-critic algorithm that characterises the actor parameter's behaviour with high probability. The bound is valid for all time after a given time instant. An all-time bound guarantees that once the algorithm enters a "safe zone" at time $n_0$, it will stay there forever with high probability. We analyse here a two-timescale actor–critic algorithm employing linear function approximation for the critic in the long-run average-reward setting and establish a lower bound on the probability that for a suitably chosen 
$n_0$ the actor parameter enters the safe region by time $n_0$
 and remains within it thereafter.

\section{Related Works}
In \cite{konda_actor_critic_type}, several actor--critic algorithms based on lookup-table representations were introduced, along with the first asymptotic convergence analysis of such methods. This line of work was extended in \cite{konda_onactorcritic}, where actor--critic algorithms employing function approximation through the Q-value function were proposed and their asymptotic convergence properties were analyzed. A natural-gradient-based actor--critic algorithm was subsequently developed in \cite{kakade_2001}. Later, works such as \cite{dicastro2009convergent} and \cite{zhang2020provably} further advanced the asymptotic analysis of actor--critic methods. In \cite{BHATNAGAR20092471}, natural actor--critic algorithms incorporating bootstrapping in both the actor and critic updates were proposed, and a comprehensive asymptotic analysis, including stability guarantees, was established. 


Over the past few years, there has been considerable interest in establishing finite-time convergence guarantees for reinforcement learning algorithms. In \cite{wu2022finite}, a finite-time analysis of a two-timescale actor--critic algorithm under Markovian sampling was carried out, yielding a sample complexity of $\tilde{\mathcal{O}}(\epsilon^{-2.5})$ for convergence to an $\epsilon$-approximate stationary point of the performance objective. Subsequently, \cite{chen2023finitetime} analyzed a single-timescale actor--critic algorithm under Markovian sampling and established a sample complexity of $\tilde{\mathcal{O}}(\epsilon^{-2})$. Finite-time convergence bounds for a natural policy gradient method applied to constrained discounted-cost Markov decision processes (MDPs) were derived in \cite{NEURIPS2020_5f7695de}. In \cite{luo2023finitetime}, the authors studied a decentralized single-timescale actor--critic algorithm and obtained a sample complexity of $\tilde{\mathcal{O}}(\epsilon^{-2})$. Furthermore, the non-asymptotic convergence behavior of a two-timescale natural actor--critic algorithm was investigated in \cite{9827586}, where a sample complexity bound of $\tilde{\mathcal{O}}(\epsilon^{-6})$ was established.  \cite{pmlr-v244-panda24a} carried out a non-asymptotic analysis of three-timescale constrained actor-critic and constrained natural actor-critic algorithms in a non-i.i.d. (Markovian) setting with a long-run average-cost criterion, where both the objective and constraint functions are suitable policy-dependent long-run averages of prescribed cost functions.

\cite{pmlr-v37-korda15} provided high-probability bounds as well as bounds in expectation for the well-known temporal difference learning algorithm TD(0) with linear function approximators. \cite{ThoppeBorkar2019} developed a concentration bound for stochastic approximation that is both tighter and valid under substantially weaker assumptions. Notably, the bound remains applicable even when the step-size sequence is not square-summable. Also, \cite{chandak2026concentrationboundtd0function} derived a uniform all-time concentration bound for TD(0) with linear function approximation.

\section{Background }

\subsection{Markov Decision Process}

We consider an MDP with finite state and action spaces that is characterised by the tuple $(S,A,P,r)$, where 
$S$ denotes the state space, 
$A$ is the action space, $P(s^{'} \vert s,a)$ is the probability of transition from state $s$ to $s^{'}$ under action $a$. Further,
$r$ denotes the random single-stage reward that depends on the state $s$ and action $a$ at the given instant and with expected value $r(s,a)$. We  let $\vert r(s,a) \vert \leq U_r$, $\forall s \in S,\forall a \in A$, where $U_r \in (0,\infty)$ is a constant.
We consider stationary randomized policies $\pi_\theta(a|s)$, $a\in A, s\in S$ parameterised by $\theta$. Our aim is to maximise the long-run average reward (with $\mu_\theta$ being the stationary distribution):
\begin{align}\label{Eq:J-function}
    L(\theta) :&= \lim\limits_{T \rightarrow \infty}\frac{1}{T}\sum\limits_{t=0}^{T-1}r(s_t,a_t) = E_{s \sim \mu_{\theta},a \sim \pi_{\theta}}[r(s,a)].
\end{align}
The differential value function $V^{\theta}(s), s\in S$ is defined as (with $s_0$ being the starting state, $a_t \sim \pi_{\theta}(\cdot|s_t)$ and $s_{t+1} \sim P(\cdot| s_t,a_t)$): ${\displaystyle 
    V^{\theta}(s) = E\bigg[\sum\limits_{t=0}^{\infty}(r(s_t,a_t) - L(\theta)) |s_0 = s \bigg]}$.
The differential action-value (Q-value) function 
is defined as
\begin{align*}
Q_{\theta}(s,a)&= \mathbb{E}_{\theta}[\sum\limits_{t=0}^{\infty}(r(s_t,a_t)-L(\theta))|s_0=s,a_0=a]\\
    &\overset{\text{(i)}}
    {=}r(s,a)-L(\theta)+\mathbb{E}[V^{\theta}(s')],
\end{align*}
where the expectation in (i) is taken over $s'\sim P(\cdot|s,a)$.

The policy gradient theorem \citep{sutton1999, suttonbarto} gives the following expression for $\nabla_\theta L(\theta)$:
\begin{align*}
\nabla_{\theta}L(\theta)=\mathbb{E}_{s\sim\mu_{\theta},a\sim\pi_{\theta}}[A_{\theta}(s,a)\nabla_{\theta}\log\pi_{\theta}(a|s)],
\end{align*}
where $A_{\theta}(s,a) =Q_\theta(s,a)-V^{\theta}(s)$ denotes the advantage function.

\subsection{Function Approximation}

In order to save on the computational effort needed to find exact solutions, one often uses value function approximation techniques based on linear or nonlinear function approximation architectures. We use linear function approximators here for our theoretical results. Such approximators have been found to be viable for asymptotic analyses. For instance, see \citep{tsitsiklisroy2} for an asymptotic analysis of temporal difference learning algorithms and \citep{BHATNAGAR20092471} for an analysis of AC algorithms when linear function approximators are used in the average cost setting.
We approximate the state-value function here using a linear approximation architecture as
${\displaystyle
\widehat{V}^{\theta}(s;v)=\phi(s)^\top v}$, 
where $\phi: \mathcal{S}\rightarrow \mathbb{R}^{d_1}$ is a known feature mapping and $\theta$ is the policy parameter for the considered policy. 

\subsection{Two-Timescale Actor-Critic Algorithm}

\begin{algorithm}[tb]
   \caption{Two Timescale Actor-Critic Algorithm}
   \label{algo}
\begin{algorithmic}
   \STATE {\bfseries Input:} initial average reward parameter $L_0$, initial actor parameter $\theta_0$, initial critic parameter $v_0$, step-size $\beta_{t}$ for actor, $\alpha_{t}$ for critic and $\gamma_{t}$ for the average reward estimator.
   \STATE Draw $s_{0}$ from some initial distribution.
   \FOR {$t=0,1,2,\dots$}
    \STATE Take the action $a_t \sim \pi_{\theta_t}(\cdot|s_t)$
    \STATE Observe next state $s_{t+1} \sim P(\cdot|s_t,a_t)$ and the reward $r_t = r(s_t,a_t)$
    \STATE $L_{t+1} = L_t + \gamma_t (r_t - L_t)$
    \STATE $\delta_t = r_t - L_t + \phi(s_{t+1})^{\top} v_{t} - \phi(s_t)^{\top} v_{t}$
    \STATE $v_{t+1} = \Gamma(v_{t} + \alpha_{t} \delta_t \phi(s_t))$\label{algline:critic_update}
    \STATE $\theta_{t+1} = \bar{\Gamma}(\theta_{t} + \beta_{t} \delta_t \nabla_{\theta} \log \pi_{\theta_{t}}(a_t|s_t))$\label{algline:actor_update}
\ENDFOR 
\end{algorithmic}
\end{algorithm}

Algorithm~\ref{algo} provides the two-timescale actor--critic (AC) algorithm, where the critic employs linear function approximation. The step-size sequences satisfy the standard Robbins--Monro conditions. Furthermore, they are chosen such that $\beta_t =o(\alpha_t)$ for all $t\geq 0$, and $\gamma_t =K \alpha_t $ for some constant $K>0$. Consequently, the average reward and critic updates evolve on the faster timescale, while the actor updates operate on the slower timescale.

To ensure the boundedness of the critic estimates, we employ the projection operator $\Gamma(\cdot)$. Specifically, for any $x\in\mathbb{R}^{d_1}$, $\Gamma(x)$ denotes the projection of $x$ onto a compact and convex set $C\subset\mathbb{R}^{d_1}$. Moreover, every vector $y\in C$ satisfies $\|y\|\leq U_v$, where $U_v>0$ is a constant. 

We also incorporate a projection operator $\bar{\Gamma}(\cdot)$ into the actor update. Specifically, for any $x\in\mathbb{R}^{d}$, $\bar{\Gamma}(x)$ denotes the projection of $x$ onto a compact, convex set $\bar{C}\subset\mathbb{R}^{d}$. Furthermore, we assume that $\bar{C}$ is bounded such that $\|y\|\leq U_\theta$ for all $y\in\bar{C}$, where $U_\theta>0$ is a fixed constant. Finally, as mentioned earlier, the single-stage reward is a function of the current state and the action selected.

\section{Assumptions and Main Result}

\subsection{Assumptions}
\begin{assumption} \label{assum:bounded_feature_norm}
    The norm of each state feature is bounded by 1,  i.e., $\Vert \phi(i) \Vert \le 1$, $\forall i\in S$.
\end{assumption}

Assumption \ref{assum:bounded_feature_norm} is essential for establishing the various upper bounds in the analysis.
\begin{assumption}
\label{assum:negative-definite}
    For all potential policy parameters $\theta$, the matrix $\Ab$ defined as under is negative definite:  
${\displaystyle
    \Ab := \EE_{s,a,s^{'}} \big[ \phi(s) \big( \phi(s^{'}) - \phi(s)\big)^{\top} \big],
}$
where $s \sim \mu_{\theta}(\cdot)$ (the stationary distribution under policy parameter $\theta$) and $a \sim \pi_{\theta}(\cdot | s), s^{'} \sim P(\cdot | s, a)$. 
Further, let $\lambda_\theta$ denote the largest eigenvalue of $\Ab$. Then $-\lambda \stackrel{\triangle}{=} \sup_\theta \lambda_\theta <0$. 
\end{assumption}
This assumption   
helps give the existence and uniqueness of $v^{*}(\theta)$ because the following equations hold: For $s \sim \mu_{\theta}(\cdot), a \sim \pi_{\theta}(\cdot | s)$,
\begin{align}
    &\Ab v^{*}(\theta) + \bbb = 0,\label{critic_conv_point}\\
    &\bbb:= \EE_{s,a,s^{'}} [(r(s,a)- L(\theta))\phi(s) ]\notag.
\end{align}

The approximation error for the feature mapping can vary depending on its complexity.
We define the approximation error that arises due to linear function approximation as follows.
\begin{align*}
    \epsilon_{\text{app}}(\btheta, s) := 
    \vert \phi(s)^{\top} v^{*}(\theta) - V^{\pi_{\btheta}}(s)\vert.
\end{align*} 
We assume that 
\begin{align*}
     \forall \btheta ,\forall s,   \mbox{ } \epsilon_{\text{app}}(\btheta ,s) \le \epsilon_{\text{app}},
\end{align*}
where  $\epsilon_{\text{app}} > 0$ is some constant.

\begin{assumption} \label{assum:policy-lipschitz-bounded}
There exist $L,B, K>0$ such that for all $s\in S$ and $a\in A$,
\begin{enumerate}
\item[(a)] $\big\|\nabla \log \pi_{\theta}(a|s) \big\| \le B$, $\forall \theta \in \RR^d$,
\item[(b)] $\big\|\nabla \log \pi_{\theta_1}(a|s) - \nabla \log \pi_{\theta_2}(a|s) \big\| \le  K \norm{\theta_1 - \theta_2}$, $\forall \theta_1,\theta_2 \in \RR^d$,
\item[(c)] $\big|\pi_{\theta_1}(a|s) - \pi_{\theta_2}(a|s) \big| \le L \norm{\theta_1 - \theta_2}$, $\forall \theta_1,\theta_2 \in \RR^d$,
\end{enumerate}
\end{assumption}

Assumptions \ref{assum:policy-lipschitz-bounded}(a) , \ref{assum:policy-lipschitz-bounded}(b), \ref{assum:policy-lipschitz-bounded}(c) are standard in the literature on policy gradient methods, see \citep{wu2022finite}. 

\begin{assumption}[Uniform Ergodicity]\label{ergodicity}
Consider a Markov chain generated as per the following: $a_t \sim \pi_{\theta}(\cdot | s_t), s_{t+1} \sim P(\cdot | s_t, a_t)$. Then there exist $b > 0$ and $k \in (0,1)$ such that:
    \begin{align*}
        d_{TV}\big(P^{\tau}(s_{\tau} \in \cdot | s_0 = s), \mu_{\theta}(\cdot)\big) \le b k^{\tau}, \forall \tau \ge 0, \forall s \in S, \forall a \in A.
    \end{align*}

Also, $ b \leq \epsilon_{\text{app}}$. In the above $d_{TV}(\mu,\nu)$ denotes the total variation distance between probability measures $\mu$ and $\nu$. 
    
\end{assumption}

\begin{assumption}[Strong Concavity of the Performance Function]
\label{concavity}
There exists a constant $m>0$ such that the  performance function $L:\RR^d \rightarrow\mathbb{R}$ is $m$-strongly concave, i.e., for any $\theta_1,\theta_2\in  \bar{C} $,
\begin{align}
\left\langle
\nabla L(\theta_1)-\nabla L(\theta_2),
\theta_1-\theta_2
\right\rangle
\leq
-m\Vert \theta_1-\theta_2\Vert^2.
\end{align}
\end{assumption}

Assumption \ref{concavity} plays a key role in establishing lemmas \ref{contraction_F} and \ref{theta_bound}.

\subsection{Concentration Bound   Results}
\begin{figure*}
    \centering
    \includegraphics[scale =0.3]{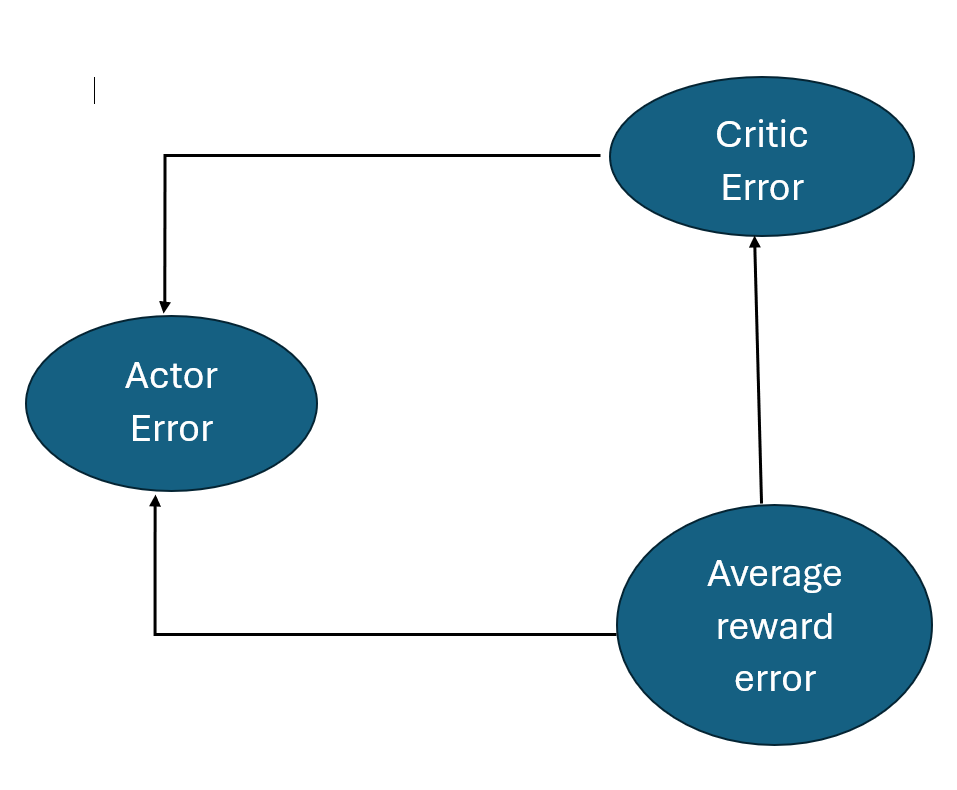}
    \caption{Dependency of errors in the analysis of actor critic algorithm for concentration bound}
    \label{fig:error_dependency}
\end{figure*}

For our analysis, we define the learning rates as follows: $\alpha_t = c_{\alpha}/(1+t)^{\nu} , \beta_{t} = c_{\beta}/(1+t)^{\sigma} , \gamma_t = c_{\gamma}/(1+t)^{\nu}$, where $c_{\alpha}, c_{\beta},c_{\gamma} $ are some positive constants and $0 < \nu < \sigma \leq 1$  , $\forall t \geq 0$. We have $\theta^{*}$ as the fixed point of $\bar{G}(\theta) = \bar{\Gamma}(\sum_{s}\sum_{a}\mu_{\theta}(s)\pi_{\theta}(a|s)F(\theta,s,a))$ where $ F(\theta,s,a) = (r(s,a)  - L(\theta) + \sum_{s^{'}}P(s^{'}|s,a)V^{\theta}(s^{'}) - V^{\theta}(s)) \nabla \log \pi_{\theta}(a|s) + \theta$. We have ${\displaystyle B < \frac{\sqrt{2m}}{L_{v}}}$ and ${\displaystyle U_r < \frac{\sqrt{2m} - L_{v}B}{\bigg( 1 + \frac{2b}{1-k} \bigg)(L_{\mu}B + BL_{\pi} + K)}}$, with constants $L_{v}, L_{\mu} , L_{\pi}>0$ defined in lemma \ref{lipschitz_mu_pi_V}. Also we select $d_1 > 0$ and $d_2 > 0$ such that $ 0 < 2\alpha + 2Bd_{1} + B d_{2} < 1$.

\begin{theorem}\label{main_result_theorem}
There exist finite positive constants $C_1, C_2,C_3,C_4, D$, such that for $0< \delta \leq 1, \varepsilon > 0$, and $n_0$ sufficiently large such that $2m\beta_{n_0} < 1,$ we have
 \begin{align*}
    & P \bigg(  \Vert \theta_m - \theta^{*} \Vert \leq    e^{-(1-\alpha)b_{n_0}(m-1)}\varepsilon +  \Delta (n_0, \delta), \forall m \geq n_0 \bigg)\\
    & \geq  1 - P \bigg(  \Vert \theta_{n_0} - \theta^{*} \Vert > \varepsilon \bigg)   -8\sum\limits_{k \geq  n_0} e^{-\frac{D\delta^4}{ \omega(k)}}.
\end{align*}
where $ \Delta (n_0, \delta) := \frac{4\delta + C_1 \sqrt{\epsilon_\text{app}} +  C_2 n_0^{(\nu - \sigma)/2} +  C_3 n_0^{-\nu /2} +  C_4 \sqrt{bk} }{1 - \sqrt{(2\alpha + 2Bd_{1} + B d_{2})}}$ and \[
\omega(m)=
\begin{cases}
(n_0)^{c_\beta - \sigma}m^{-c_\beta}, & \text{if } c_\beta \leq \sigma, \\
 m ^{-\sigma},  & otherwise.
\end{cases}
\]

\end{theorem}

As illustrated in Figure \ref{fig:error_dependency}, the actor error analysis depends on both the critic error and the error in the estimated average reward. Hence, we also analyzed the average reward estimation error and the critic error to establish the results presented in Theorem \ref{main_result_theorem}.

\begin{corollary}\label{corollary}

For $\sigma = 1$, $\nu = 0.5$, $\varepsilon \geq 1$ and sufficiently large $n_0$ such that $2m\beta_{n_0} < 1,$
    \begin{align*}
   &P\bigg( \Vert \theta_k - \theta^{*} \Vert = O \bigg( \frac{n_0^{3/4}}{k} \frac{1}{\sqrt{\epsilon_2}}+   \bigg(\frac{1}{n_0}\bigg)^{1/4} \log^{1/4} \bigg( \frac{1}{\epsilon_1  }\bigg)   + \bigg(\frac{1}{n_0} \bigg)^{1/4} \bigg) \bigg)\\
    &\qquad \geq 1 - \epsilon_1 - \epsilon_2,
\end{align*}

$\forall  k \geq n_0.$
\end{corollary}

We provide below a proof sketch of Theorem~\ref{main_result_theorem} and Corollary~\ref{corollary}. The proof begins by formulating the actor update as a stochastic approximation iteration. We then define an iterative sequence $\{z_n\}_{n\geq n_0}$ and use the decomposition

$$
\|\theta_n-\theta^*\|
\leq
\|\theta_n-z_n\|+\|z_n-\theta^*\|.
$$

Next, we derive probabilistic bounds for $\Vert\theta_n-z_n\Vert$ and $\Vert z_n-\theta^*\Vert$. Combining these bounds yields the desired probabilistic bound on $\Vert \theta_n-\theta^*\Vert$.

\section{Experimental Results}

We may mention that the actor-critic algorithm is very well studied in the literature for its empirical properties and performance comparisons with other related algorithms. However, in order to
empirically validate the convergence of  Algorithm~\ref{algo}, we evaluate it on three different Open AI Gym environments, namely CartPole-v1, FrozenLake-v1, and Blackjack-v1, respectively, each reformulated as a continuing average-reward task by resetting the environment state to an initial state upon the termination of an episode. In other words, once the environment hits the goal state, rather than treating the latter as absorbing, we randomly pick another non-terminal state for a restart and continue.

Since the optimal parameter $\theta^{*}$ has no closed-form expression in these settings, we approximate it by running Algorithm~\ref{algo} for a large number of iterations across several independent seeds and averaging the resulting parameters. Figure  \ref{fig:plots} plots $\Vert\theta_t - \theta^{*}\Vert$ against the iteration count $t$, averaged over ten independent training runs with a shaded band indicating one standard deviation. It can be seen that in all three environments, the actor parameter exhibits a smooth, monotonically decreasing trend, consistent with the theoretical two-timescale convergence guarantee. The hyperparameters used for these experiments are illustrated in Table~\ref{tab:actor-critic-arch}. 

\section{Conclusions and Future Work}
To the best of our knowledge, we obtained the first finite-time, high-probability convergence bound for the two-timescale actor-critic algorithm. Our bound characterises the behaviour of the actor parameter with high probability. The bound holds uniformly (over time instants) after a specified time instant. An interesting direction for future work would be to establish analogous concentration bounds for other classes of actor-critic algorithms, such as the two-timescale natural actor-critic and the three-timescale constrained actor-critic algorithms. This would provide insights into how learning rates, constraints and natural gradients affect the concentration bound.

\begin{figure*}
    \centering
    \includegraphics[scale =0.1]{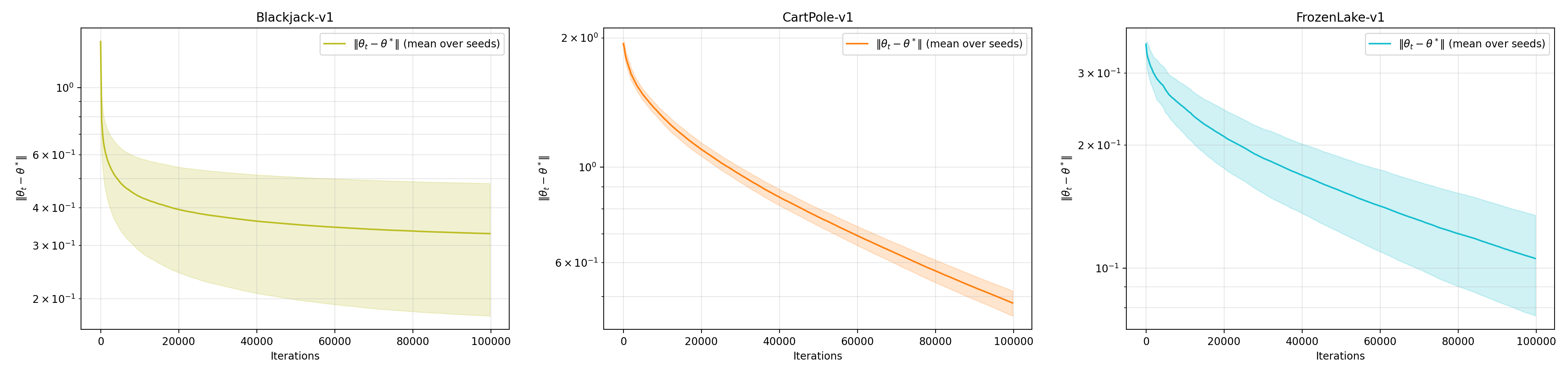}
    \caption{ Convergence of $\Vert\theta_t - \theta^{*}\Vert$   averaged over ten independent runs}
    \label{fig:plots}
\end{figure*}

\bibliography{iclr2027_conference}

@article{konda_actor_critic_type,
author = {Konda, Vijay and Borkar, Vivek},
year = {1999},
month = {12},
pages = {94-123},
title = {Actor-Critic--Type Learning Algorithms for Markov Decision Processes},
volume = {38},
journal = {SIAM J. Control and Optimization},
doi = {10.1137/S036301299731669X}
}

@article{konda_onactorcritic,
author = {Konda, Vijay R. and Tsitsiklis, John N.},
title = {OnActor-Critic Algorithms},
journal = {SIAM Journal on Control and Optimization},
volume = {42},
number = {4},
pages = {1143-1166},
year = {2003},
doi = {10.1137/S0363012901385691},

URL = { 
    
        https://doi.org/10.1137/S0363012901385691
    
    
},
eprint = { 
    
        https://doi.org/10.1137/S0363012901385691
    
    

}
}

@inproceedings{kakade_2001,
 author = {Kakade, Sham M},
 booktitle = {Advances in Neural Information Processing Systems},
 editor = {T. Dietterich and S. Becker and Z. Ghahramani},
 pages = {},
 publisher = {MIT Press},
 title = {A Natural Policy Gradient},
 url = {https://proceedings.neurips.cc/paper_files/paper/2001/file/4b86abe48d358ecf194c56c69108433e-Paper.pdf},
 volume = {14},
 year = {2001}
}

@misc{dicastro2009convergent,
      title={A Convergent Online Single Time Scale Actor Critic Algorithm}, 
      author={D. Di Castro and R. Meir},
      year={2009},
      eprint={0909.2934},
      archivePrefix={arXiv},
      primaryClass={cs.LG}
}

@misc{zhang2020provably,
      title={Provably Convergent Two-Timescale Off-Policy Actor-Critic with Function Approximation}, 
      author={Shangtong Zhang and Bo Liu and Hengshuai Yao and Shimon Whiteson},
      year={2020},
      eprint={1911.04384},
      archivePrefix={arXiv},
      primaryClass={cs.LG}
}

@article{BHATNAGAR20092471,
title = {Natural actor–critic algorithms},
journal = {Automatica},
volume = {45},
number = {11},
pages = {2471-2482},
year = {2009},
issn = {0005-1098},
doi = {https://doi.org/10.1016/j.automatica.2009.07.008},
url = {https://www.sciencedirect.com/science/article/pii/S0005109809003549},
author = {Shalabh Bhatnagar and Richard S. Sutton and Mohammad Ghavamzadeh and Mark Lee},
}

@article{bhatnagar2023actorcritic,
  title={Actor-Critic or Critic-Actor? A Tale of Two Time Scales},
  author={Bhatnagar, Shalabh and Borkar, Vivek S and Guin, Soumyajit},
  journal={IEEE Control Systems Letters},
  year={2023},
  volume={7},
pages = {2671-2676},
  publisher={IEEE}
}

@misc{panda_bhatnagar_2025_CA,
      title={Two-Timescale Critic-Actor for Average Reward MDPs with Function Approximation}, 
      author={Prashansa Panda and Shalabh Bhatnagar},
      year={2025},
      eprint={2402.01371},
      archivePrefix={arXiv},
      primaryClass={cs.LG},
      url={https://arxiv.org/abs/2402.01371}, 
}

@misc{wu2022finite,
     title={A Finite Time Analysis of Two Time-Scale Actor Critic Methods}, 
      author={Yue Wu and Weitong Zhang and Pan Xu and Quanquan Gu},
     year={2022},
     eprint={2005.01350},
     archivePrefix={arXiv},
      primaryClass={cs.LG}
}

@misc{chen2023finitetime,
      title={Finite-time analysis of single-timescale actor-critic}, 
      author={Xuyang Chen and Lin Zhao},
      year={2023},
      eprint={2210.09921},
      archivePrefix={arXiv},
      primaryClass={cs.LG}
}

@inproceedings{NEURIPS2020_5f7695de,
 author = {Ding, Dongsheng and Zhang, Kaiqing and Basar, Tamer and Jovanovic, Mihailo},
 booktitle = {Advances in Neural Information Processing Systems},
 pages = {8378--8390},
 publisher = {Curran Associates, Inc.},
 title = {Natural Policy Gradient Primal-Dual Method for Constrained Markov Decision Processes},
 url = {https://proceedings.neurips.cc/paper_files/paper/2020/file/5f7695debd8cde8db5abcb9f161b49ea-Paper.pdf},
 volume = {33},
 year = {2020}
}

@misc{luo2023finitetime,
      title={Finite-Time Analysis of Fully Decentralized Single-Timescale Actor-Critic}, 
      author={Qijun Luo and Xiao Li},
      year={2023},
      eprint={2206.05733},
      archivePrefix={arXiv},
      primaryClass={cs.LG}
}

@ARTICLE{9827586,
  author={Khodadadian, Sajad and Doan, Thinh T. and Romberg, Justin and Maguluri, Siva Theja},
  journal={IEEE Transactions on Automatic Control}, 
  title={Finite-Sample Analysis of Two-Time-Scale Natural Actor–Critic Algorithm}, 
  year={2023},
  volume={68},
  number={6},
  pages={3273-3284},
  doi={10.1109/TAC.2022.3190032}}

@article{ThoppeBorkar2019,
  author    = {Gugan Thoppe and Vivek S. Borkar},
  title     = {A Concentration Bound for Stochastic Approximation via Alekseev's Formula},
  journal   = {Stochastic Systems},
  volume     = {9},
  number     = {1},
  pages      = {1--26},
  year       = {2019},
  doi        = {10.1287/stsy.2018.0019},
  publisher  = {INFORMS}
}

@article{chandak2026concentrationboundtd0function,
  author  = {Chandak, Siddharth and Borkar, Vivek S.},
  title   = {A Concentration Bound for {TD}(0) with Function Approximation},
  journal = {Stochastic Systems},
  volume  = {16},
  number  = {1},
  pages   = {44--60},
  year    = {2026},
  doi     = {10.1287/stsy.2023.0055},
  url     = {https://doi.org/10.1287/stsy.2023.0055}
}

@article{sutton1999,
  title={Policy gradient methods for reinforcement learning with function approximation},
  author={Sutton, R.S. and McAllester, D. and Singh, S. and Mansour, Y.},
  journal={Advances in neural information processing systems},
  volume={12},
  year={1999}
}

@book{suttonbarto,
  title={Reinforcement learning: An introduction},
  author={Sutton, R.S. and Barto, A.G.},
  year={2018},
  publisher={MIT press}
}

@InProceedings{pmlr-v244-panda24a,
  title = 	 {Finite-Time Analysis of Three-Timescale Constrained Actor-Critic and Constrained Natural Actor-Critic Algorithms.},
  author =       {Panda, Prashansa and Bhatnagar, Shalabh},
  booktitle = 	 {Proceedings of the Fortieth Conference on Uncertainty in Artificial Intelligence},
  pages = 	 {2787--2834},
  year = 	 {2024},
  editor = 	 {Kiyavash, Negar and Mooij, Joris M.},
  volume = 	 {244},
  series = 	 {Proceedings of Machine Learning Research},
  month = 	 {15--19 Jul},
  publisher =    {PMLR},
  url = 	 {https://proceedings.mlr.press/v244/panda24a.html}
}

@InProceedings{pmlr-v37-korda15,
  title = 	 {On TD(0) with function approximation: Concentration bounds and a centered variant with exponential convergence},
  author = 	 {Korda, Nathaniel and La, Prashanth},
  booktitle = 	 {Proceedings of the 32nd International Conference on Machine Learning},
  pages = 	 {626--634},
  year = 	 {2015},
  editor = 	 {Bach, Francis and Blei, David},
  volume = 	 {37},
  series = 	 {Proceedings of Machine Learning Research},
  address = 	 {Lille, France},
  month = 	 {07--09 Jul},
  publisher =    {PMLR},
  url = 	 {https://proceedings.mlr.press/v37/korda15.html}
}

@book{Bertsekas2013OptimalControlVol2,
  author    = {Bertsekas, Dimitri P.},
  title     = {Optimal Control and Dynamic Programming},
  volume    = {2},
  edition   = {4},
  year      = {2013},
  publisher = {Athena Scientific},
  address   = {Belmont, MA}
}

@article{tsitsiklisroy2,
  title={Average cost temporal-difference learning},
  author={Tsitsiklis, J.N. and Van Roy, B.},
  journal={Automatica},
  volume={35},
  number={11},
  pages={1799--1808},
  year={1999},
  publisher={Elsevier}
}
\bibliographystyle{iclr2027_conference}

\newpage

\appendix
\section{Appendix}

\subsection{Proof of Main Result}

For $n,m \geq 0$, let $\xi (n,m) = \prod_{k=m}^{k=n} (1-\beta_k)$ if $n \geq m$ and 1 otherwise.\\

There are several lemmas that will be used in the proof.

\begin{lemma}\label{lipschitz_L}
There exists a constant $M_{L} > 0$ such that for all $ \theta_1 , \theta_2 \in \RR^d,$ it holds that
 \begin{align*}
     \Vert L(\theta_1) - L(\theta_2)\Vert \leq M_{L}\Vert \theta_1 - \theta_2 \Vert .
 \end{align*}   
\end{lemma}

\begin{lemma}\label{lipschitz_mu_pi_V}

There exists $ L_{v} > 0 , L_{\mu} > 0, L_{\pi} > 0$ such that such that for all $ \theta_1 , \theta_2 \in \RR^d, s \in S,$ 
\begin{align*}
    \sum_{s}\vert\mu_{\theta_1}(s) - \mu_{\theta_2}(s)\vert &\leq L_{\mu}\Vert \theta_1 - \theta_2 \Vert,\\
    \Vert V^{\theta_1}(s) - V^{\theta_2}(s)  \Vert &\leq L_{v}\Vert \theta_1 - \theta_2 \Vert,\\
    \sum_{a}\vert \pi_{\theta_1}(a|s) - \pi_{\theta_2}(a|s) \vert &\leq L_{\pi}\Vert \theta_1 - \theta_2 \Vert.
\end{align*}
    
\end{lemma}

\begin{lemma}\label{average_reward_convergence}
 \begin{align*}
     &\sum_{k = n_0}^{k = m}\xi(m,k+1)\beta_k((L_k - L(\theta_k))^2  \leq  \mathcal{O}(n_0^{\nu - \sigma})  + \mathcal{O}(n_0^{-\nu }) + \mathcal{O}(bk)\\
    &\qquad + \sum_{k = n_0}^{k = m}\xi(m,k+1)\beta_k (L_k - L(\theta_k)) \bigg( r(s_k,a_k) - \sum_{a}\pi_{\theta_k}(a|s_k)r(s_k,a) + \bar{r}_{\theta_k}(s_{k+1}) -\sum_{s }P^{\theta_k}(s |s_k)\bar{r}_{\theta_k}(s)\bigg)
 \end{align*}  
 where $\bar{r}_{\theta}(s) = \sum_{a}\pi_{\theta}(a|s)r(s,a) $ and $ P^{\theta_k}(s |s_k) = \sum\limits_{a}P(s|s_k,a)\pi_{\theta_k}(a|s_k)$.
\end{lemma}

\begin{lemma}\label{critic_convergence}
\begin{align*}
    &\sum_{k = n_0}^{k = m}\xi(m,k+1)\beta_k\Vert v_k - v^{*}(\theta_k) \Vert^2 \leq  \mathcal{O}(n_0^{\nu - \sigma}) +  \mathcal{O}(n_0^{-\nu}) + \mathcal{O}(bk)\\
    &\qquad +  \frac{2}{\lambda}\sum_{k = n_0}^{k = m}\xi(m,k+1)\beta_k \langle z_k , W(s_k,a_k,s_{k+1}, L_k,v_k)  - \sum\limits_{a}\sum\limits_{s^{'}}\pi_{\theta_k}(a|s_k)P(s^{'}|s_k,a)W(s_k,a,s^{'}, L_k,v_k)\rangle \\
     &\qquad + \frac{2}{\lambda}\sum_{k = n_0}^{k = m}\xi(m,k+1)\beta_k \langle z_k , \bar{W}_{\theta_k}(s_{k+1},L_k , v_k) - \sum_{s} P^{\theta_k}(s|s_k) \bar{W}_{\theta_k}(s,L_k , v_k) \rangle\\
    &\qquad + \frac{1}{2 \lambda}\sum_{k = n_0}^{k = m}\xi(m,k+1)\beta_k y_k \bigg( r(s_k,a_k) - \sum_{a}\pi_{\theta_k}(a|s_k)r(s_k,a) + \bar{r}_{\theta_k}(s_{k+1}) -\sum_{s }P^{\theta_k}(s |s_k)\bar{r}_{\theta_k}(s)\bigg) 
\end{align*}

where $W(s,a,s^{'}, L,v) = (r(s,a) - L + \phi(s^{'})^T v - \phi(s)^Tv)\phi(s)$ , $\bar{W}_{\theta}(s,L,v) = \sum\limits_{a}\sum\limits_{s^{'}}\pi_{\theta_k}(a|s)P(s^{'}|s,a)W(s,a,s^{'}, L,v)$ and $z_k = v_k - v^{*}(\theta_k).$
    
\end{lemma}

Now we will formulate the actor update as a stochastic approximation iteration.

We have the following update for actor:-

\begin{align*}
    \theta_{t+1} &= \theta_{t} + \beta_{t} \delta_t \nabla \log \pi_{\theta_t}(a_t|s_t)\\
    & = \theta_{t} + \beta_{t} (r(s_t,a_t)  - L_t +  \phi(s_{t+1})^{\top} v_{t} - \phi(s_t)^{\top} v_{t}) \nabla \log \pi_{\theta_t}(a|s)\\
    &= \theta_{t} + \beta_{t}(r(s_t,a_t)  - L_t + V^{\theta_t}(s_{t+1}) - V^{\theta_t}(s_{t})) \nabla \log \pi_{\theta_t}(a|s)\\
    &\qquad + \beta_t (\phi(s_{t+1})^{\top} v_{t} -  V^{\theta_t}(s_{t+1}) - \phi(s_t)^{\top} v_{t} + V^{\theta_t}(s_{t}) ) \nabla \log \pi_{\theta_t}(a|s)\\
    &= \theta_{t} + \beta_{t}(r(s_t,a_t) - L(\theta_t) +\sum_{s^{'}}P(s^{'}|s_t,a_t)V^{\theta_t}(s^{'}) - V^{\theta_t}(s_{t})) \nabla \log \pi_{\theta_t}(a|s)\\
    &\qquad + \beta_t (\phi(s_{t+1})^{\top} v_{t} -  V^{\theta_t}(s_{t+1}) - \phi(s_t)^{\top} v_{t} + V^{\theta_t}(s_{t}) ) \nabla \log \pi_{\theta_t}(a_t|s_t)\\
    &\qquad +  \beta_t(V^{\theta_t}(s_{t+1}) - \sum_{s^{'}}P(s^{'}|s_t,a_t)V^{\theta_t}(s^{'}))\nabla \log \pi_{\theta_t}(a_t|s_t)\\
     &\qquad + \beta_t(L_t - L(\theta_t)) \nabla \log \pi_{\theta_t}(a_t|s_t)\\
    &= \theta_{t} + \beta_{t}(F(\theta_t,s_t,a_t) - \theta_t) + \beta_t M_{t+1}\\
    &\qquad +
    \beta_t (\phi(s_{t+1})^{\top} v_{t} -  V^{\theta_t}(s_{t+1}) - \phi(s_t)^{\top} v_{t} + V^{\theta_t}(s_{t}) ) \nabla \log \pi_{\theta_t}(a_t|s_t)\\
     &\qquad + \beta_t(L_t - L(\theta_t)) \nabla \log \pi_{\theta_t}(a_t|s_t)\\
    &= \theta_{t} + \beta_{t}(\sum_{s}\sum_{a}\mu_{\theta_t}(s)\pi_{\theta_t}(a|s)F(\theta_t,s,a) - \theta_t) + \underbrace{\beta_t M_{t+1}}_{I_1}\\
    &\qquad +
    \beta_t (\phi(s_{t+1})^{\top} v_{t} - V^{\theta_t}(s_{t+1}) - \phi(s_t)^{\top} v_{t} + V^{\theta_t}(s_{t}) ) \nabla \log \pi_{\theta_t}(a_t|s_t)\\
    &\qquad + \underbrace{\beta_t (F(\theta_t,s_t,a_t) - \sum_{s}\sum_{a}\mu_{\theta_t}(s)\pi_{\theta_t}(a|s)F(\theta_t,s,a))}_{I_2}\\
    &\qquad +\beta_t(L_t - L(\theta_t)) \nabla \log \pi_{\theta_t}(a_t|s_t)\\
\end{align*}
where 
\begin{align*}
    F(\theta,s,a) = (r(s,a)  - L(\theta) + \sum_{s^{'}}P(s^{'}|s,a)V^{\theta}(s^{'}) - V^{\theta}(s)) \nabla \log \pi_{\theta}(a|s) + \theta, 
\end{align*}
and
\begin{align*}
    M_{t+1} =  (V^{\theta_t}(s_{t+1}) - \sum_{s^{'}}P(s^{'}|s_t,a_t)V^{\theta_t}(s^{'}))\nabla \log \pi_{\theta_t}(a_t|s_t)
\end{align*}

Define the family of $\sigma-$ fields $\textit{F}_{n} := \sigma (\theta_{m}, s_{m}, a_m, m \leq n ), n \geq 0$. Then $\{M_{n}\}$ is a martingale difference sequence with respect to ${\textit{F}_{n}};$ that is,

\begin{align*}
    E[M_{n+1}|\textit{F}_{n} ] = 0, a.s. \forall n,
\end{align*}

The term $I_1$ denotes the error because of the martingale noise term and term $I_2$ denotes the error because of Markov noise.

\begin{lemma}\label{contraction_F}

For any $\theta_1, \theta_2 \in \RR^d, $
\begin{align*}
    \bigg\Vert  \sum_{s}\sum_{a}\mu_{\theta_1}(s)\pi_{\theta_1}(a|s)F(\theta_1,s,a)  - \sum_{s}\sum_{a}\mu_{\theta_2}(s)\pi_{\theta_2}(a|s)F(\theta_2,s,a)\bigg\Vert \leq \alpha \Vert \theta_1 - \theta_2 \Vert
\end{align*}

where $0 < \alpha < 1$.
\end{lemma}

Define $z_n$ for $n \geq n_0$ by
\begin{align*}
    z_{n+1} = \bar{\Gamma}\bigg(z_n + \beta_{n}\bigg(\sum_{s , a }\mu_{z_{n}}(s)\pi_{z_{n}}(a|s)F(z_n,s,a) - z_n \bigg)\bigg)
\end{align*}
where $z_{n_0} = \theta_{n_0}$.

We assume $2m\beta_{n_0} < 1$. Note that $\Vert \theta_n - \theta^{*}\Vert \leq \Vert \theta_n - z_n\Vert + \Vert  z_n - \theta^{*}\Vert$.\\
Now let us bound the second term. 

Recall that the Euclidean projection onto a nonempty,
closed, convex set $\bar{C}$ satisfies the variational
characterization
\begin{equation}
    y=\bar{\Gamma}(x)
    \quad\Longleftrightarrow\quad
    \langle x-y,z-y\rangle\leq 0,
    \qquad \forall z\in\bar{C}.\notag
\end{equation}

Since
\begin{equation}
    \theta^{*}
    =
    \bar{\Gamma}\left(
        \theta^{*}+\nabla L(\theta^{*})
    \right),\notag
\end{equation}
the variational characterization implies that
\begin{equation}
    \left\langle
        \nabla L(\theta^{*}),
        \theta-\theta^{*}
    \right\rangle
    \leq 0,
    \qquad \forall\theta\in\bar{C}.\notag
\end{equation}

Multiplying both sides by $\beta_{n}>0$, we obtain
\begin{equation}
    \left\langle
        \beta_{n}\nabla L(\theta^{*}),
        \theta-\theta^{*}
    \right\rangle
    \leq 0,
    \qquad \forall\theta\in\bar{C}.\notag
\end{equation}

Equivalently,
\begin{equation}
    \left\langle
        \left(
            \theta^{*}
            +\beta_{n}\nabla L(\theta^{*})
        \right)-\theta^{*},
        \theta-\theta^{*}
    \right\rangle
    \leq 0,
    \qquad \forall\theta\in\bar{C}.\notag
\end{equation}

Applying the variational characterization of
Euclidean projection once again yields
\begin{equation}
    \boxed{
    \theta^{*}
    =
    \bar{\Gamma}\left(
        \theta^{*}
        +\beta_{n}\nabla L(\theta^{*})
    \right).
    }
\end{equation}

Since $\beta_{n}>0$ was arbitrary, the identity holds
for every positive actor step size. Hence we have,

\begin{align}
    & \Vert  z_{n+1} - \theta^{*} \Vert \notag\\
    &=  \bigg\Vert\bar{\Gamma}\bigg( z_n + \beta_{n}\bigg(\sum_{s}\sum_{a}\mu_{z_{n}}(s)\pi_{z_{n}}(a|s)F(z_n,s,a) - z_n \bigg)\bigg) - \theta^{*} \bigg \Vert \notag \\
    &= \bigg\Vert\bar{\Gamma}( z_n + \beta_{n} \nabla L(z_n))- \theta^{*} \bigg \Vert \notag\\
    &= \bigg\Vert\bar{\Gamma}( z_n + \beta_{n} \nabla L(z_n))- \bar{\Gamma}(  \theta^{*} + \beta_{n} \nabla L(\theta^{*})) \bigg \Vert \notag\\
    &\leq  \bigg\Vert z_n + \beta_{n} \nabla L(z_n)-   \theta^{*} - \beta_{n} \nabla L(\theta^{*}) \bigg \Vert \notag\\
    &= \left\|
(1-\beta_n)(z_n-\theta^*)
+
\beta_n(G(z_n)-G(\theta^*))
\right\|
\nonumber \\
&\leq
\left(
1-(1-\alpha)\beta_n
\right)
\|z_n-\theta^*\| \notag\\
  &\leq e^{-(1-\alpha)b_{n_0}(n)}\Vert z_{n_0} - \theta^{*} \Vert \label{first_ineq}
\end{align}

where $b_{n_0}(n) = \sum\limits_{k = n_0}^{n}\beta_{k}$.

Next we give a probabilistic bound on the term $\Vert \theta_n - z_n\Vert$. We have ,

\begin{align*}
    &\Vert \theta_{t+1} - z_{t+1} \Vert^2\\
    &= \Vert \bar{\Gamma}(\theta_{t} + \beta_{t}(\sum_{s}\sum_{a}\mu_{\theta_t}(s)\pi_{\theta_t}(a|s)F(\theta_t,s,a) - \theta_t) + \beta_t M_{t+1}\\
    &\qquad +
    \beta_t (\phi(s_{t+1})^{\top} v_{t} -  V^{\theta_t}(s_{t+1}) - \phi(s_t)^{\top} v_{t} + V^{\theta_t}(s_{t}) ) \nabla \log \pi_{\theta_t}(a|s)\\
    &\qquad + \beta_t (F(\theta_t,s_t,a_t) - \sum_{s}\sum_{a}\mu_{\theta_t}(s)\pi_{\theta_t}(a|s)F(\theta_t,s,a))\\
    &\qquad + \beta_t(L_t - L(\theta_t)) \nabla \log \pi_{\theta_t}(a_t|s_t))\\
    &\qquad - \bar{\Gamma}\bigg(z_t + \beta_{t}\bigg(\sum_{s , a }\mu_{z_{t}}(s)\pi_{z_{t}}(a|s)F(z_t,s,a) - z_t \bigg)\bigg)\Vert^2\\
    &\leq  \Vert(1 - \beta_t)(\theta_t - z_t) + \beta_t M_{t+1} + \beta_t (F(\theta_t,s_t,a_t) -\sum_{s , a }\mu_{z_{t}}(s)\pi_{z_{t}}(a|s)F(z_t,s,a) )\\
    &\qquad + \beta_t (\phi(s_{t+1})^{\top} v_{t} -  V^{\theta_t}(s_{t+1}) - \phi(s_t)^{\top} v_{t} + V^{\theta_t}(s_{t}) ) \nabla \log \pi_{\theta_t}(a|s)\\
    &\qquad +\beta_t(L_t - L(\theta_t)) \nabla \log \pi_{\theta_t}(a_t|s_t)\Vert^2\\
    &= \Vert (1 - \beta_t)(\theta_t - z_t) + \beta_t M_{t+1} + \beta_t (\sum_{s , a }\mu_{\theta_{t}}(s)\pi_{\theta_{t}}(a|s)F(\theta_t,s,a) -\sum_{s , a }\mu_{z_{t}}(s)\pi_{z_{t}}(a|s)F(z_t,s,a) )\\
    &\qquad + \beta_t (\phi(s_{t+1})^{\top} v_{t} -  V^{\theta_t}(s_{t+1}) - \phi(s_t)^{\top} v_{t} + V^{\theta_t}(s_{t}) ) \nabla \log \pi_{\theta_t}(a|s)\\
    &\qquad + \beta_t(F(\theta_t,s_t,a_t) -\sum_{s , a }\mu_{\theta_{t}}(s)\pi_{\theta_{t}}(a|s)F(\theta_t,s,a) )\\
    &\qquad + \beta_t(L_t - L(\theta_t)) \nabla \log \pi_{\theta_t}(a_t|s_t)\Vert^2\\
    &= \Vert (1 - \beta_t)(\theta_t - z_t) + \beta_t M_{t+1} + \beta_t (\sum_{s , a }\mu_{\theta_{t}}(s)\pi_{\theta_{t}}(a|s)F(\theta_t,s,a) -\sum_{s , a }\mu_{z_{t}}(s)\pi_{z_{t}}(a|s)F(z_t,s,a) )\\
    &\qquad + \beta_t(F(\theta_t,s_t,a_t) -\sum_{s , a }\mu_{\theta_{t}}(s)\pi_{\theta_{t}}(a|s)F(\theta_t,s,a) )\\
     &\qquad + \beta_t (\phi(s_{t+1})^{\top} v_{t} -  \phi(s_{t+1})^{\top} v^{*}(\theta_t) - \phi(s_t)^{\top} v_{t} + \phi(s_t)^{\top} v^{*}(\theta_t) ) \nabla \log \pi_{\theta_t}(a|s)\\
     &\qquad + \beta_t ( \phi(s_{t+1})^{\top} v^{*}(\theta_t) -  V^{\theta_t}(s_{t+1}) - \phi(s_{t})^{\top} v^{*}(\theta_t) + V^{\theta_t}(s_{t}))  \nabla \log \pi_{\theta_t}(a|s)\\
     &\qquad + \beta_t(L_t - L(\theta_t)) \nabla \log \pi_{\theta_t}(a_t|s_t)\Vert^2.
\end{align*}

Let us denote 
\begin{align*}
    D_t &= M_{t+1} + (\sum_{s , a }\mu_{\theta_{t}}(s)\pi_{\theta_{t}}(a|s)F(\theta_t,s,a) -\sum_{s , a }\mu_{z_{t}}(s)\pi_{z_{t}}(a|s)F(z_t,s,a) )\\
    &\qquad + (F(\theta_t,s_t,a_t) -\sum_{s , a }\mu_{\theta_{t}}(s)\pi_{\theta_{t}}(a|s)F(\theta_t,s,a) )\\
     &\qquad +  (\phi(s_{t+1})^{\top} v_{t} -  \phi(s_{t+1})^{\top} v^{*}(\theta_t) - \phi(s_t)^{\top} v_{t} + \phi(s_t)^{\top} v^{*}(\theta_t) ) \nabla \log \pi_{\theta_t}(a|s)\\
     &\qquad +  ( \phi(s_{t+1})^{\top} v^{*}(\theta_t) -  V^{\theta_t}(s_{t+1}) - \phi(s_{t})^{\top} v^{*}(\theta_t) + V^{\theta_t}(s_{t}))  \nabla \log \pi_{\theta_t}(a|s)\\
     &\qquad + (L_t - L(\theta_t)) \nabla \log \pi_{\theta_t}(a_t|s_t)\Vert^2.
\end{align*}

Hence we have,
\begin{align*}
    &\Vert \theta_{t+1} - z_{t+1} \Vert^2\\
    \leq & (1-\beta_t)\Vert \theta_t - z_t\Vert^2 + \beta_t^2 D_t^2 + 2 \beta_{t}(1 - \beta_t) \langle \theta_t - z_t , D_t \rangle 
\end{align*}

 For some $ n \geq n_0$ we iterate the above for $n_0 \leq m \leq n $ to obtain

\begin{align}
    &\Vert \theta_{m+1} - z_{m+1}\Vert^2\notag\\
    &= \sum_{k = n_0}^{k = m}\xi(m,k+1)\beta_k^2 D_{k}^2  + 2 \bigg\vert\sum_{k = n_0}^{k = m}\xi(m,k+1) \beta_{k}(1 - \beta_{k}) \langle \theta_k - z_k , M_{k+1} \rangle \bigg\vert\notag\\
    & +  2\sum_{k = n_0}^{k = m}\xi(m,k+1) \beta_{k}(1 - \beta_{k}) \langle \theta_k - z_k , \sum_{s , a }\mu_{\theta_{k}}(s)\pi_{\theta_{k}}(a|s)F(\theta_k,s,a) -\sum_{s , a }\mu_{z_{k}}(s)\pi_{z_{k}}(a|s)F(z_k,s,a) \rangle  \notag\\
    & +  \underbrace{ 2\sum_{k = n_0}^{k = m}\xi(m,k+1) \beta_{k}(1 - \beta_{k}) \langle \theta_k - z_k , F(\theta_k,s_k,a_k) -\sum_{s , a }\mu_{\theta_{k}}(s)\pi_{\theta_{k}}(a|s)F(\theta_k,s,a) \rangle}_{I_1} \notag\\
    &+  \underbrace{ 2\sum_{k = n_0}^{k = m}\xi(m,k+1) \beta_{k}(1 - \beta_{k}) \langle \theta_k - z_k ,(\phi(s_{k+1})^{\top} v_{k} -  \phi(s_{k+1})^{\top} v^{*}(\theta_k) - \phi(s_k)^{\top} v_{k} + \phi(s_k)^{\top} v^{*}(\theta_k) ) \nabla \log \pi_{\theta_k}(a_k|s_k)\rangle}_{I_2} \notag\\
    &+ \underbrace{ 2\sum_{k = n_0}^{k = m}\xi(m,k+1) \beta_{k}(1 - \beta_{k}) \langle \theta_k - z_k , ( \phi(s_{k+1})^{\top} v^{*}(\theta_k) -  V^{\theta_k}(s_{k+1}) - \phi(s_{k})^{\top} v^{*}(\theta_k) + V^{\theta_k}(s_{k}))  \nabla \log \pi_{\theta_k}(a_k|s_k)\rangle}_{I_3}\notag\\
    &+ \underbrace{ 2\sum_{k = n_0}^{k = m}\xi(m,k+1) \beta_{k}(1 - \beta_{k}) \langle \theta_k - z_k ,((L_k - L(\theta_k)) \nabla \log \pi_{\theta_k}(a_k|s_k)\rangle}_{I_4} \label{ineq_second_bound}
\end{align}

Here we use the definition that $z_{n_0} = \theta_{n_0}$. 

For any $0 < k \leq m$,
\begin{align*}
  \xi(m,k)  +  \xi(m,k+1)\beta_k = \xi(m, k+1),
\end{align*}

and hence
\begin{align*}
    \xi(m,n_0) +  \sum\limits_{k = n_0}^{m}\xi(m,k+1)\beta_k = \xi(m, m+1) = 1.
\end{align*}

which implies 
\begin{align*}
    \sum\limits_{k = n_0}^{m}\xi(m,k+1)\beta_k \leq 1
\end{align*}

For term $I_1$ we have ,
\begin{align*}
    I_1 &= 2\sum_{k = n_0}^{k = m}\xi(m,k+1) \beta_{k}(1 - \beta_{k}) \langle \theta_k - z_k , F(\theta_k,s_k,a_k) -\sum_{s , a }\mu_{\theta_{k}}(s)\pi_{\theta_{k}}(a|s)F(\theta_k,s,a) \rangle\\
    &= 2\sum_{k = n_0}^{k = m}\xi(m,k+1) \beta_{k}(1 - \beta_{k}) \langle \theta_k - z_k , U(\theta_k,s_k,a_k) -\sum_{s , a }\mu_{\theta_{k}}(s)\pi_{\theta_{k}}(a|s)U(\theta_k,s,a) \rangle\\
    &= 2\sum_{k = n_0}^{k = m}\xi(m,k+1) \beta_{k}(1 - \beta_{k}) \langle \theta_k - z_k ,( U(\theta_k,s_k,a_k) - \sum_{a}\pi_{\theta_k}(a|s_k)U(\theta_k,s_k,a))\rangle\\
    &\qquad +2\sum_{k = n_0}^{k = m}\xi(m,k+1) \beta_{k}(1 - \beta_{k}) \langle \theta_k - z_k ,( \sum_{a}\pi_{\theta_k}(a|s_k)U(\theta_k,s_k,a) - \sum_{s , a }\mu_{\theta_{k}}(s)\pi_{\theta_{k}}(a|s)U(\theta_k,s,a) )\rangle\\
    &= 2\sum_{k = n_0}^{k = m}\xi(m,k+1) \beta_{k}(1 - \beta_{k}) \langle \theta_k - z_k ,\bigg( U(\theta_k,s_k,a_k) - \sum_{a}\pi_{\theta_k}(a|s_k)U(\theta_k,s_k,a)\bigg)\rangle\\
    &\qquad +2\sum_{k = n_0}^{k = m}\xi(m,k+1) \beta_{k}(1 - \beta_{k}) \langle \theta_k - z_k ,(\bar{U}(\theta_k,s_k) -\sum_{s }P^{\theta_k}(s |s_k)\bar{U}(\theta_k,s) )\rangle\\
    &\qquad + 2\sum_{k = n_0}^{k = m}\xi(m,k+1) \beta_{k}(1 - \beta_{k}) \langle \theta_k - z_k ,\bigg(\sum_{s }(P^{\theta_k}(s |s_k) - \mu_{\theta_k}(s) )\bar{U}(\theta_k,s)\bigg)\rangle\\
     &= 2\sum_{k = n_0}^{k = m}\xi(m,k+1) \beta_{k}(1 - \beta_{k}) \langle \theta_k - z_k ,\bigg( U(\theta_k,s_k,a_k) - \sum_{a}\pi_{\theta_k}(a|s_k)U(\theta_k,s_k,a)\bigg)\rangle\\
    &\qquad +2\sum_{k = n_0}^{k = m}\xi(m,k+1) \beta_{k}(1 - \beta_{k}) \langle \theta_k - z_k ,(\bar{U}(\theta_k,s_{k+1}) -\sum_{s }P^{\theta_k}(s |s_k)\bar{U}(\theta_k,s) )\rangle\\
    &\qquad + \underbrace{2\sum_{k = n_0}^{k = m}\xi(m,k+1) \beta_{k}(1 - \beta_{k}) \langle \theta_k - z_k ,(\bar{U}(\theta_k,s_{k}) - \bar{U}(\theta_k,s_{k+1}))\rangle}_{I_{1a}}\\
    &\qquad + \underbrace{2\sum_{k = n_0}^{k = m}\xi(m,k+1) \beta_{k}(1 - \beta_{k}) \langle \theta_k - z_k ,\bigg(\sum_{s }(P^{\theta_k}(s |s_k) - \mu_{\theta_k}(s) )\bar{U}(\theta_k,s)\bigg)\rangle}_{I_{1b}}\\
\end{align*}

where we define
\begin{align*}
    &U(\theta, s ,a) = F(\theta, s, a) - \theta,\\
    & \bar{U}(\theta, s) = \sum_{a}\pi_{\theta}(a|s)U(\theta,s,a),\\
    & P^{\theta_k}(s |s_k) = \sum\limits_{a}P(s|s_k,a)\pi_{\theta_k}(a|s_k).
\end{align*}

Now for term $I_{1a}$ we have,

\begin{align*}
    &2\sum_{k = n_0}^{k = m}\xi(m,k+1) \beta_{k}(1 - \beta_{k}) \langle \theta_k - z_k ,(\bar{U}(\theta_k,s_{k}) - \bar{U}(\theta_k,s_{k+1}))\rangle\\
    =& 2\sum_{k = n_0}^{k = m}\xi(m,k+1) \beta_{k}(1 - \beta_{k}) \langle \theta_k - z_k ,(\bar{U}(\theta_k,s_{k}) - \bar{U}(\theta_{k+1},s_{k+1}))\rangle \\
    &\qquad + 2\sum_{k = n_0}^{k = m}\xi(m,k+1) \beta_{k}(1 - \beta_{k}) \langle \theta_k - z_k , (\bar{U}(\theta_{k+1},s_{k+1}) -\bar{U}(\theta_k,s_{k+1}) )\rangle\\
    =& 2\sum_{k = n_0}^{k = m}\xi(m,k+1) \beta_{k}(1 - \beta_{k}) \langle \theta_k - z_k , \bar{U}(\theta_k,s_{k})\rangle -  2\sum_{k = n_0}^{k = m}\xi(m,k+1) \beta_{k}(1 - \beta_{k}) \langle \theta_k - z_k , \bar{U}(\theta_{k+1},s_{k+1})\rangle\\
    &\qquad + 2\sum_{k = n_0}^{k = m}\xi(m,k+1) \beta_{k}(1 - \beta_{k}) \langle \theta_k - z_k , (\bar{U}(\theta_{k+1},s_{k+1}) -\bar{U}(\theta_k,s_{k+1}) )\rangle\\
    =& 2\sum_{k = n_0}^{k = m}\xi(m,k+1) \beta_{k}(1 - \beta_{k}) \langle \theta_k - z_k , \bar{U}(\theta_k,s_{k}) \rangle-  2\sum_{k = n_0 + 1}^{k = m + 1}\xi(m,k)\beta_{k-1}(1 - \beta_{k-1}) \langle \theta_{k-1} -z_{k-1}, \bar{U}(\theta_{k},s_{k})\rangle\\
    &\qquad + 2\sum_{k = n_0}^{k = m}\xi(m,k+1) \beta_{k}(1 - \beta_{k}) \langle \theta_k - z_k , (\bar{U}(\theta_{k+1},s_{k+1}) -\bar{U}(\theta_k,s_{k+1}) )\rangle\\
    =&2\sum_{k = n_0}^{k = m}\xi(m,k+1) \beta_{k}(1 - \beta_{k}) \langle \theta_k - z_k , \bar{U}(\theta_k,s_{k})\rangle -  2\sum_{k = n_0 }^{k = m }\xi(m,k)\beta_{k-1}(1 - \beta_{k-1})\langle \theta_{k-1}- z_{k-1},  \bar{U}(\theta_{k},s_{k})\rangle\\
    &\qquad + 2\xi(m,n_0)\beta_{n_0-1}( 1- \beta_{n_0 - 1}) \langle \theta_{n_0 - 1} - z_{n_0 - 1},\bar{U}(\theta_{n_0},s_{n_0})\rangle \\
    &\qquad - 2\xi(m,m+1)\beta_{m}( 1- \beta_{m})\langle \theta_m - z_m ,  \bar{U}(\theta_{m+1},s_{m+1}) \rangle\\
    &\qquad + 2\sum_{k = n_0}^{k = m}\xi(m,k+1) \beta_{k}(1 - \beta_{k}) \langle \theta_k - z_k , (\bar{U}(\theta_{k+1},s_{k+1}) -\bar{U}(\theta_k,s_{k+1}) )\rangle\\
     =& 2\sum_{k = n_0}^{k = m} \langle (\xi(m,k+1) \beta_{k}(1 - \beta_{k}))(\theta_k - z_k)-  (\xi(m,k) \beta_{k-1}(1 - \beta_{k-1}))(\theta_{k-1} - z_{k-1}) , \bar{U}(\theta_k,s_{k})\rangle \\
   &\qquad + 2\xi(m,n_0)\beta_{n_0-1}( 1- \beta_{n_0 - 1}) \langle \theta_{n_0 - 1} - z_{n_0 - 1},\bar{U}(\theta_{n_0},s_{n_0})\rangle \\
    &\qquad - 2\xi(m,m+1)\beta_{m}( 1- \beta_{m})\langle \theta_m - z_m ,  \bar{U}(\theta_{m+1},s_{m+1}) \rangle\\
    &\qquad + 2\sum_{k = n_0}^{k = m}\xi(m,k+1) \beta_{k}(1 - \beta_{k}) \langle \theta_k - z_k , (\bar{U}(\theta_{k+1},s_{k+1}) -\bar{U}(\theta_k,s_{k+1}) )\rangle\\
\end{align*}

Hence we have ,
\begin{align*}
     I_{1a}  &=  \mathcal{O}(n_0^{ -\nu})
\end{align*}

For term $I_{1b}$ we have,
\begin{align*}
   &2\sum_{k = n_0}^{k = m}\xi(m,k+1) \beta_{k}(1 - \beta_{k}) \langle \theta_k - z_k ,\bigg(\sum_{s }(P^{\theta_k}(s |s_k) - \mu_{\theta_k}(s) )\bar{U}(\theta_k,s)\bigg)\rangle \\
   =& O(bk)
\end{align*}

Hence we have ,
\begin{align*}
    I_1 &=  2\sum_{k = n_0}^{k = m}\xi(m,k+1) \beta_{k}(1 - \beta_{k}) \langle \theta_k - z_k ,\bigg( U(\theta_k,s_k,a_k) - \sum_{a}\pi_{\theta_k}(a|s_k)U(\theta_k,s_k,a)\bigg)\rangle\\
    &\qquad +2\sum_{k = n_0}^{k = m}\xi(m,k+1) \beta_{k}(1 - \beta_{k}) \langle \theta_k - z_k ,(\bar{U}(\theta_k,s_{k+1}) -\sum_{s }P^{\theta_k}(s |s_k)\bar{U}(\theta_k,s) )\rangle\\
    &\qquad +  \mathcal{O}(n_0^{ -\nu}) +  \mathcal{O}(bk)
\end{align*}

For term $I_2$ we have,
\begin{align*}
    &2\sum_{k = n_0}^{k = m}\xi(m,k+1) \beta_{k}(1 - \beta_{k}) \langle \theta_k - z_k ,(\phi(s_{k+1})^{\top} v_{k} -  \phi(s_{k+1})^{\top} v^{*}(\theta_k) - \phi(s_k)^{\top} v_{k} + \phi(s_k)^{\top} v^{*}(\theta_k) ) \nabla \log \pi_{\theta_k}(a_k|s_k)\rangle\\
    &\leq 4B\sum_{k = n_0}^{k = m}\xi(m,k+1) \beta_{k}\Vert \theta_k - z_k \Vert \Vert  v_k - v^{*}(\theta_k)\Vert \\
\end{align*}

Now we have, for any $d_1 > 0$,

\begin{align*}
    &2 \Vert \theta_k - z_k \Vert \Vert  v_k - v^{*}(\theta_k)\Vert\\
    \leq & d_1 \Vert \theta_k - z_k \Vert^2 + \frac{1}{d_1}\Vert  v_k - v^{*}(\theta_k)\Vert^2
\end{align*}

Hence we have,
\begin{align*}
    I_2 & \leq 2B d_1\sum_{k = n_0}^{k = m}\xi(m,k+1) \beta_{k} \Vert \theta_k - z_k \Vert^2 + \frac{2B}{d_1}\sum_{k = n_0}^{k = m}\xi(m,k+1) \beta_{k}\Vert  v_k - v^{*}(\theta_k)\Vert^2
\end{align*}

For term $I_3$ we have,

\begin{align*}
& 2\sum_{k = n_0}^{k = m}\xi(m,k+1) \beta_{k}(1 - \beta_{k}) \langle \theta_k - z_k , ( \phi(s_{k+1})^{\top} v^{*}(\theta_k) -  V^{\theta_k}(s_{k+1}) - \phi(s_{k})^{\top} v^{*}(\theta_k) + V^{\theta_k}(s_{k}))  \nabla \log \pi_{\theta_k}(a_k|s_k)\rangle \\& = \mathcal{O}(\epsilon_\text{app})
\end{align*}

For term $I_4$ we have ,
\begin{align*}
    & 2\sum_{k = n_0}^{k = m}\xi(m,k+1) \beta_{k}(1 - \beta_{k}) \langle \theta_k - z_k ,((L_k - L(\theta_k)) \nabla \log \pi_{\theta_k}(a_k|s_k)\rangle\\
    \leq & 2 B\sum_{k = n_0}^{k = m}\xi(m,k+1)\beta_k \Vert \theta_k - z_k \Vert \vert L_k - L(\theta_k)\vert\\
\end{align*}

Now we have, for any $d_2 > 0$,
\begin{align*}
    2\Vert \theta_k - z_k \Vert \vert L_k - L(\theta_k)\vert \leq d_2 \Vert \theta_k - z_k \Vert^2 + \frac{1}{d_2}\vert L_k - L(\theta_k)\vert ^2
\end{align*}
 Hence  we have,

 \begin{align*}
     I_4 &= B d_{2}\sum_{k = n_0}^{k = m}\xi(m,k+1)\beta_k \Vert \theta_k - z_k \Vert^2 + \frac{B}{d_2}\sum_{k = n_0}^{k = m}\xi(m,k+1)\beta_k \vert L_k - L(\theta_k)\vert ^2
 \end{align*}

Hence after gathering all the terms we have ,

\begin{align*}
    &\Vert \theta_{m+1} - z_{m+1} \Vert ^2 \\
    &\leq \sum_{k = n_0}^{k = m}\xi(m,k+1)\beta_k^2 D_{k}^2  + 2 \bigg\vert\sum_{k = n_0}^{k = m}\xi(m,k+1) \beta_{k}(1 - \beta_{k}) \langle \theta_k - z_k , M_{k+1} \rangle \bigg\vert\notag\\
    & +  2\sum_{k = n_0}^{k = m}\xi(m,k+1) \beta_{k}(1 - \beta_{k}) \langle \theta_k - z_k , \sum_{s , a }\mu_{\theta_{k}}(s)\pi_{\theta_{k}}(a|s)F(\theta_k,s,a) -\sum_{s , a }\mu_{z_{k}}(s)\pi_{z_{k}}(a|s)F(z_k,s,a) \rangle \\
    &\qquad +  2\sum_{k = n_0}^{k = m}\xi(m,k+1) \beta_{k}(1 - \beta_{k}) \langle \theta_k - z_k ,\bigg( U(\theta_k,s_k,a_k) - \sum_{a}\pi_{\theta_k}(a|s_k)U(\theta_k,s_k,a)\bigg)\rangle\\
    &\qquad +2\sum_{k = n_0}^{k = m}\xi(m,k+1) \beta_{k}(1 - \beta_{k}) \langle \theta_k - z_k ,(\bar{U}(\theta_k,s_{k+1}) -\sum_{s }P^{\theta_k}(s |s_k)\bar{U}(\theta_k,s) )\rangle\\
    &\qquad +  \mathcal{O}(n_0^{ -\nu}) +  \mathcal{O}(bk) + \mathcal{O}(\epsilon_\text{app})\\
    &\qquad + 2B d_1\sum_{k = n_0}^{k = m}\xi(m,k+1) \beta_{k} \Vert \theta_k - z_k \Vert^2 + \frac{2B}{d_1}\sum_{k = n_0}^{k = m}\xi(m,k+1) \beta_{k}\Vert  v_k - v^{*}(\theta_k)\Vert^2\\
    &\qquad + B d_{2}\sum_{k = n_0}^{k = m}\xi(m,k+1)\beta_k \Vert \theta_k - z_k \Vert^2 + \frac{B}{d_2}\sum_{k = n_0}^{k = m}\xi(m,k+1)\beta_k \vert L_k - L(\theta_k)\vert ^2\\
    &\leq 2 \bigg\vert\sum_{k = n_0}^{k = m}\xi(m,k+1) \beta_{k}(1 - \beta_{k}) \langle \theta_k - z_k , M_{k+1} \rangle \bigg\vert\\
    &\qquad + (2\alpha + 2Bd_{1} + B D_{2})  \sum_{k = n_0}^{k = m}\xi(m,k+1)\beta_k \Vert \theta_k - z_k\Vert^2\\
    &\qquad + 2\sum_{k = n_0}^{k = m}\xi(m,k+1) \beta_{k}(1 - \beta_{k}) \langle \theta_k - z_k ,\bigg( U(\theta_k,s_k,a_k) - \sum_{a}\pi_{\theta_k}(a|s_k)U(\theta_k,s_k,a)\bigg)\rangle\\
    &\qquad +2\sum_{k = n_0}^{k = m}\xi(m,k+1) \beta_{k}(1 - \beta_{k}) \langle \theta_k - z_k ,(\bar{U}(\theta_k,s_{k+1}) -\sum_{s }P^{\theta_k}(s |s_k)\bar{U}(\theta_k,s) )\rangle\\
    &\qquad +  \mathcal{O}(n_0^{ -\nu}) +  \mathcal{O}(bk) + \mathcal{O}(\epsilon_\text{app})\\
    &\qquad + \frac{2B}{d_1}\sum_{k = n_0}^{k = m}\xi(m,k+1) \beta_{k}\Vert  v_k - v^{*}(\theta_k)\Vert^2+ \frac{B}{d_2}\sum_{k = n_0}^{k = m}\xi(m,k+1)\beta_k \vert L_k - L(\theta_k)\vert ^2\\
\end{align*}

Now using the result of lemmas \ref{average_reward_convergence} and \ref{critic_convergence} we have ,

\begin{align*}
     &\Vert \theta_{m+1} - z_{m+1} \Vert^2\\
     &\leq 2 \bigg\vert\sum_{k = n_0}^{k = m}\xi(m,k+1) \beta_{k}(1 - \beta_{k}) \langle \theta_k - z_k , M_{k+1} \rangle \bigg\vert\\
    &\qquad + (2\alpha + 2Bd_{1} + B D_{2})  \sum_{k = n_0}^{k = m}\xi(m,k+1)\beta_k \Vert \theta_k - z_k\Vert^2\\
    &\qquad +  2\sum_{k = n_0}^{k = m}\xi(m,k+1) \beta_{k}(1 - \beta_{k}) \langle \theta_k - z_k ,\bigg( U(\theta_k,s_k,a_k) - \sum_{a}\pi_{\theta_k}(a|s_k)U(\theta_k,s_k,a)\bigg)\rangle\\
    &\qquad +2\sum_{k = n_0}^{k = m}\xi(m,k+1) \beta_{k}(1 - \beta_{k}) \langle \theta_k - z_k ,(\bar{U}(\theta_k,s_{k+1}) -\sum_{s }P^{\theta_k}(s |s_k)\bar{U}(\theta_k,s) )\rangle\\
    &\qquad +  \mathcal{O}(n_0^{ -\nu}) +  \mathcal{O}(bk) + \mathcal{O}(\epsilon_\text{app})\\
    &\qquad + \mathcal{O}(n_0^{\nu - \sigma})  + \mathcal{O}(n_0^{-\nu })\\
    &\qquad +  \frac{B}{d_2}\sum_{k = n_0}^{k = m}\xi(m,k+1)\beta_k (L_k - L(\theta_k)) \bigg( r(s_k,a_k) - \sum_{a}\pi_{\theta_k}(a|s_k)r(s_k,a) + \bar{r}_{\theta_k}(s_{k+1}) -\sum_{s }P^{\theta_k}(s |s_k)\bar{r}_{\theta_k}(s)\bigg)\\
    &\qquad + \frac{4B}{d_1\lambda}\sum_{k = n_0}^{k = m}\xi(m,k+1)\beta_k \langle z_k , W(s_k,a_k,s_{k+1}, L_k,v_k)  - \sum\limits_{a}\sum\limits_{s^{'}}\pi_{\theta_k}(a|s_k)P(s^{'}|s_k,a)W(s_k,a,s^{'}, L_k,v_k)\rangle \\
     &\qquad + \frac{4B}{d_1\lambda}\sum_{k = n_0}^{k = m}\xi(m,k+1)\beta_k \langle z_k , \bar{W}_{\theta_k}(s_{k+1},L_k , v_k) - \sum_{s} P^{\theta_k}(s|s_k) \bar{W}_{\theta_k}(s,L_k , v_k) \rangle\\
    &\qquad + \frac{B}{d_1 \lambda}\sum_{k = n_0}^{k = m}\xi(m,k+1)\beta_k y_k \bigg( r(s_k,a_k) - \sum_{a}\pi_{\theta_k}(a|s_k)r(s_k,a) + \bar{r}_{\theta_k}(s_{k+1}) -\sum_{s }P^{\theta_k}(s |s_k)\bar{r}_{\theta_k}(s)\bigg) 
\end{align*}

Let 
\begin{align*}
    \tilde{U}_{k+1} &= \langle \theta_k - z_k ,\bigg( U(\theta_k,s_k,a_k) - \sum_{a}\pi_{\theta_k}(a|s_k)U(\theta_k,s_k,a) + \bar{U}(\theta_k,s_{k+1}) -\sum_{s }P^{\theta_k}(s |s_k)\bar{U}(\theta_k,s)\bigg)\rangle\\
    \tilde{W}_{k+1} &= W(s_k,a_k,s_{k+1}, L_k,v_k)  - \sum\limits_{a}\sum\limits_{s^{'}}\pi_{\theta_k}(a|s_k)P(s^{'}|s_k,a)W(s_k,a,s^{'}, L_k,v_k)\\
    &\qquad + \bar{W}_{\theta_k}(s_{k+1},L_k , v_k) - \sum_{s} P^{\theta_k}(s|s_k) \bar{W}_{\theta_k}(s,L_k , v_k)\\ 
    \tilde{Z}_{k+1} &= r(s_k,a_k) - \sum_{a}\pi_{\theta_k}(a|s_k)r(s_k,a) + \bar{r}_{\theta_k}(s_{k+1}) -\sum_{s }P^{\theta_k}(s |s_k)\bar{r}_{\theta_k}(s)
\end{align*}

We define the family of $\sigma$- fields : $\mathcal{A}_{n} := \sigma(s_0, \theta_m,L_{m}, v_{m}, s_m , a_{m-1} , 1\leq m \leq n)$. Then $\{ \tilde{W}_{n}\} , \{\tilde{U}_{n} \} , \{\tilde{Z}_{n} \}$ are martingale   difference
sequences, with respect to $\{ \mathcal{A}_{n} \}$.\\

Hence we  have,
\begin{align*}
     &\Vert \theta_{m+1} - z_{m+1} \Vert^2\\
      &\leq 2 \bigg\vert\sum_{k = n_0}^{k = m}\xi(m,k+1) \beta_{k}(1 - \beta_{k}) \langle \theta_k - z_k , M_{k+1} \rangle \bigg\vert+ (2\alpha + 2Bd_{1} + B D_{2})  \sum_{k = n_0}^{k = m}\xi(m,k+1)\beta_k \Vert \theta_k - z_k\Vert^2\\
      &\qquad + 2\sum_{k = n_0}^{k = m}\xi(m,k+1) \beta_{k}(1 - \beta_{k}) \tilde{U}_{k+1}\\
      &\qquad +  \mathcal{O}(n_0^{ -\nu}) +  \mathcal{O}(bk) + \mathcal{O}(\epsilon_\text{app}) + \mathcal{O}(n_0^{\nu - \sigma})  \\
       &\qquad +  \bigg(\frac{B}{d_2} + \frac{B}{d_1 \lambda}\bigg)\bigg\vert\sum_{k = n_0}^{k = m}\xi(m,k+1)\beta_k y_{k}  \tilde{Z}_{k+1}\bigg\vert\\
    &\qquad + \frac{4B}{d_1\lambda}\bigg\vert\sum_{k = n_0}^{k = m}\xi(m,k+1)\beta_k \langle z_k , \tilde{W}_{k+1}\rangle \bigg\vert \\
\end{align*}

Now let us define $\theta_m^{'} = \sup\limits_{n_0 \leq k \leq m}\Vert \theta_m - z_m \Vert$. Hence we have ,

\begin{align}
     &\Vert \theta_{m+1} - z_{m+1} \Vert^2 \notag \\
      &\leq 2 \bigg\vert\sum_{k = n_0}^{k = m}\xi(m,k+1) \beta_{k}(1 - \beta_{k}) \langle \theta_k - z_k , M_{k+1} \rangle \bigg\vert   + (2\alpha + 2Bd_{1} + B D_{2})  (\theta_{m}^{'} )^2\notag\\
      &\qquad + 2\sum_{k = n_0}^{k = m}\xi(m,k+1) \beta_{k}(1 - \beta_{k}) \tilde{U}_{k+1}  +  \mathcal{O}(n_0^{ -\nu}) +  \mathcal{O}(bk) + \mathcal{O}(\epsilon_\text{app}) + \mathcal{O}(n_0^{\nu - \sigma}) \notag \\
       &\qquad +  \bigg(\frac{B}{d_2} + \frac{B}{d_1 \lambda}\bigg)\bigg\vert\sum_{k = n_0}^{k = m}\xi(m,k+1)\beta_k y_{k}  \tilde{Z}_{k+1}\bigg\vert \notag \\
    &\qquad + \frac{4B}{d_1\lambda}\bigg\vert\sum_{k = n_0}^{k = m}\xi(m,k+1)\beta_k \langle z_k , \tilde{W}_{k+1}\rangle \bigg\vert \notag \\
    \Rightarrow & \Vert \theta_{m+1} - z_{m+1} \Vert \notag \\
    &\leq \sqrt{2 \bigg\vert\sum_{k = n_0}^{k = m}\xi(m,k+1) \beta_{k}(1 - \beta_{k}) \langle \theta_k - z_k , M_{k+1} \rangle \bigg\vert }  + \sqrt{(2\alpha + 2Bd_{1} + B D_{2})}  \theta_{m}^{'} \notag\\
      &\qquad + \sqrt{2 \bigg\vert \sum_{k = n_0}^{k = m}\xi(m,k+1) \beta_{k}(1 - \beta_{k}) \tilde{U}_{k+1}}\bigg\vert  +  \mathcal{O}(n_0^{ -\nu/2}) +  \mathcal{O}(\sqrt{bk}) + \mathcal{O}(\sqrt{\epsilon_\text{app}}) + \mathcal{O}(n_0^{(\nu - \sigma)/2}) \notag \\
       &\qquad +  \sqrt{\bigg(\frac{B}{d_2} + \frac{B}{d_1 \lambda}\bigg)\bigg\vert\sum_{k = n_0}^{k = m}\xi(m,k+1)\beta_k y_{k}  \tilde{Z}_{k+1}\bigg\vert} \notag \\
    &\qquad + \sqrt{\frac{4B}{d_1\lambda}\bigg\vert\sum_{k = n_0}^{k = m}\xi(m,k+1)\beta_k \langle z_k , \tilde{W}_{k+1}\rangle \bigg\vert}\label{ineq_theta_m} 
\end{align}

Now we have ,

\begin{align*}
   &\bigg\{\sqrt{2 \bigg\vert\sum_{k = n_0}^{k = m}\xi(m,k+1) \beta_{k}(1 - \beta_{k}) \langle \theta_k - z_k , M_{k+1} \rangle \bigg\vert } \leq \delta \bigg\} \\
    &\qquad \bigcap \bigg\{ \sqrt{2 \bigg\vert \sum_{k = n_0}^{k = m}\xi(m,k+1) \beta_{k}(1 - \beta_{k}) \tilde{U}_{k+1}}\bigg\vert \leq \delta\bigg\}\\
    &\qquad \bigcap \bigg\{ \sqrt{\bigg(\frac{B}{d_2} + \frac{B}{d_1 \lambda}\bigg)\bigg\vert\sum_{k = n_0}^{k = m}\xi(m,k+1)\beta_k y_{k}  \tilde{Z}_{k+1}\bigg\vert} \leq \delta\bigg\} \bigcap \bigg\{\sqrt{\frac{4B}{d_1\lambda}\bigg\vert\sum_{k = n_0}^{k = m}\xi(m,k+1)\beta_k \langle z_k , \tilde{W}_{k+1}\rangle \bigg\vert } \leq \delta \bigg\}\\
    &\qquad \bigcap \bigg\{\theta_{m}^{'} \leq \Delta(\delta , n_0) \bigg\}\\
    & \subseteq  \bigg\{ \Vert \theta_{m+1} - z_{m+1} \Vert \leq \Delta(\delta , n_0) \bigg\}\\
    \Rightarrow  &\bigg\{\sqrt{2 \bigg\vert\sum_{k = n_0}^{k = m}\xi(m,k+1) \beta_{k}(1 - \beta_{k}) \langle \theta_k - z_k , M_{k+1} \rangle \bigg\vert } \leq \delta \bigg\} \\
    &\qquad \bigcap \bigg\{ \sqrt{2 \bigg\vert \sum_{k = n_0}^{k = m}\xi(m,k+1) \beta_{k}(1 - \beta_{k}) \tilde{U}_{k+1}}\bigg\vert \leq \delta\bigg\}\\
    &\qquad \bigcap \bigg\{ \sqrt{\bigg(\frac{B}{d_2} + \frac{B}{d_1 \lambda}\bigg)\bigg\vert\sum_{k = n_0}^{k = m}\xi(m,k+1)\beta_k y_{k}  \tilde{Z}_{k+1}\bigg\vert} \leq \delta\bigg\} \bigcap \bigg\{\sqrt{\frac{4B}{d_1\lambda}\bigg\vert\sum_{k = n_0}^{k = m}\xi(m,k+1)\beta_k \langle z_k , \tilde{W}_{k+1}\rangle \bigg\vert } \leq \delta \bigg\}\\
    &\qquad \bigcap \bigg\{\theta_{m}^{'} \leq \Delta(\delta , n_0) \bigg\}\\
    & \subseteq  \bigg\{ \Vert \theta_{m+1} - z_{m+1} \Vert \leq \Delta(\delta , n_0) \bigg\}\bigcap \bigg\{\theta_{m}^{'} \leq \Delta(\delta , n_0) \bigg\}\\
     \Rightarrow  &\bigg\{\sqrt{2 \bigg\vert\sum_{k = n_0}^{k = m}\xi(m,k+1) \beta_{k}(1 - \beta_{k}) \langle \theta_k - z_k , M_{k+1} \rangle \bigg\vert } \leq \delta \bigg\} \\
    &\qquad \bigcap \bigg\{ \sqrt{2 \bigg\vert \sum_{k = n_0}^{k = m}\xi(m,k+1) \beta_{k}(1 - \beta_{k}) \tilde{U}_{k+1}}\bigg\vert \leq \delta\bigg\}\\
    &\qquad \bigcap \bigg\{ \sqrt{\bigg(\frac{B}{d_2} + \frac{B}{d_1 \lambda}\bigg)\bigg\vert\sum_{k = n_0}^{k = m}\xi(m,k+1)\beta_k y_{k}  \tilde{Z}_{k+1}\bigg\vert} \leq \delta\bigg\} \bigcap \bigg\{\sqrt{\frac{4B}{d_1\lambda}\bigg\vert\sum_{k = n_0}^{k = m}\xi(m,k+1)\beta_k \langle z_k , \tilde{W}_{k+1}\rangle \bigg\vert } \leq \delta \bigg\}\\
    &\qquad \bigcap \bigg\{\theta_{m}^{'} \leq \Delta(\delta , n_0) \bigg\}\\
    & \subseteq  \bigg\{\theta_{m + 1}^{'} \leq \Delta(\delta , n_0) \bigg\}\\
\end{align*}

Hence we have,
\begin{align}
  & P \bigg(\bigg\{\sqrt{2 \bigg\vert\sum_{k = n_0}^{k = m}\xi(m,k+1) \beta_{k}(1 - \beta_{k}) \langle \theta_k - z_k , M_{k+1} \rangle \bigg\vert } >  \delta \bigg\}\bigg) \notag \\
    &\qquad + P \bigg( \bigg\{ \sqrt{2 \bigg\vert \sum_{k = n_0}^{k = m}\xi(m,k+1) \beta_{k}(1 - \beta_{k}) \tilde{U}_{k+1}}\bigg\vert > \delta\bigg\}\bigg) \notag\\
    &\qquad  + P \bigg( \bigg\{ \sqrt{\bigg(\frac{B}{d_2} + \frac{B}{d_1 \lambda}\bigg)\bigg\vert\sum_{k = n_0}^{k = m}\xi(m,k+1)\beta_k y_{k}  \tilde{Z}_{k+1}\bigg\vert} > \delta\bigg\}\bigg) + P \bigg( \bigg\{\sqrt{\frac{4B}{d_1\lambda}\bigg\vert\sum_{k = n_0}^{k = m}\xi(m,k+1)\beta_k \langle z_k , \tilde{W}_{k+1}\rangle \bigg\vert } >  \delta \bigg\}\bigg)\notag\\
    &\qquad + P \bigg( \bigg\{\theta_{m}^{'} >\Delta(\delta , n_0) \bigg\}\bigg)\notag\\
    & \geq   P \bigg(\theta_{m + 1}^{'} > \Delta(\delta , n_0) \bigg)\label{inequality_v_m} 
\end{align}

Now we will obtain a bound on the probability :
\begin{align*}
    P\bigg( \Vert \theta_m - \theta^{*}\Vert \leq e^{-(1-\alpha)b_{n_0}(m-1)}\varepsilon + \Delta (n_0, \delta), \forall  m \geq n_0 \bigg),
\end{align*}

for some $\varepsilon > 0, \delta >0 $. We define ,
\begin{align*}
     \Delta (n_0, \delta) := \frac{4\delta + C_1 \sqrt{\epsilon_\text{app}} +  C_2 n_0^{(\nu - \sigma)/2} +  C_3 n_0^{-\nu /2} +  C_4 \sqrt{bk} }{1 - \sqrt{(2\alpha + 2Bd_{1} + B d_{2})}}
\end{align*}

where $C_1, C_2, C_3, C_4$ are positive constants.

From (\ref{first_ineq}) we have ,
\begin{align*}
    \Vert z_{n+1} - \theta^{*} \Vert \leq e^{-(1-\alpha)b_{n_0}(n)}\Vert \theta_{n_0} - \theta^{*} \Vert
\end{align*}

and hence,

\begin{align*}
    \Vert \theta_{n_0} - \theta^{*} \Vert \leq \varepsilon  \implies  \Vert z_{n} - \theta^{*} \Vert \leq e^{-(1-\alpha)b_{n_0}(n-1)}\varepsilon.
\end{align*}

Also recall that $\theta_m^{'} = \sup\limits_{n_0 \leq k \leq m}\Vert \theta_k - z_k \Vert$. Hence

\begin{align*}
    &\{ \Vert \theta_{n_0} - \theta^{*} \Vert \leq \varepsilon    \}  \bigcap \{ \theta_m^{'} \leq  \Delta (n_0, \delta) \}\\
     & \subseteq  \{ \Vert \theta_m - \theta^{*} \Vert \leq    e^{-(1-\alpha)b_{n_0}(m-1)}\varepsilon +  \Delta (n_0, \delta), \forall m \geq n_0  \}
\end{align*}
This implies the following relation between the probabilities of the two sets.

\begin{align}
    & P \bigg(  \Vert \theta_m - \theta^{*} \Vert \leq    e^{-(1-\alpha)b_{n_0}(m-1)}\varepsilon +  \Delta (n_0, \delta), \forall m \geq n_0 \bigg) \notag\\
    & \geq 1 - P \bigg( \{  \Vert \theta_{n_0} - \theta^{*} \Vert > \varepsilon    \}  \bigcup \{ \theta_m^{'} >  \Delta (n_0, \delta)  \}  \bigg). \label{final_ineq}
\end{align}

Let us define 
\begin{align*}
    p_{m} &=  P \bigg(\bigg\{\sqrt{2 \bigg\vert\sum_{k = n_0}^{k = m}\xi(m,k+1) \beta_{k}(1 - \beta_{k}) \langle \theta_k - z_k , M_{k+1} \rangle \bigg\vert } >  \delta \bigg\}\bigg) \\
    &\qquad + P \bigg( \bigg\{ \sqrt{2 \bigg\vert \sum_{k = n_0}^{k = m}\xi(m,k+1) \beta_{k}(1 - \beta_{k}) \tilde{U}_{k+1}}\bigg\vert > \delta\bigg\}\bigg)\\
    &\qquad  + P \bigg( \bigg\{ \sqrt{\bigg(\frac{B}{d_2} + \frac{B}{d_1 \lambda}\bigg)\bigg\vert\sum_{k = n_0}^{k = m}\xi(m,k+1)\beta_k y_{k}  \tilde{Z}_{k+1}\bigg\vert} > \delta\bigg\}\bigg) + P \bigg( \bigg\{\sqrt{\frac{4B}{d_1\lambda}\bigg\vert\sum_{k = n_0}^{k = m}\xi(m,k+1)\beta_k \langle z_k , \tilde{W}_{k+1}\rangle \bigg\vert } >  \delta \bigg\}\bigg)
\end{align*}

Hence we have from (\ref{inequality_v_m}),
\begin{align*}
    P(\theta_m^{'} > \Delta (n_0, \delta)) \leq p_{m-1} +  P(  \theta_{m-1}^{'} >  \Delta (n_0, \delta))
\end{align*}

Iterating this we get 
\begin{align*}
   P(\theta_m^{'} > \Delta (n_0, \delta)) \leq \sum\limits_{k = n_0 }^{ m-1} p_{k} 
\end{align*}

Now , going back to (\ref{final_ineq}) we have,

\begin{align*}
    & P \bigg(  \Vert \theta_m - \theta^{*} \Vert \leq    e^{-(1-\alpha)b_{n_0}(m-1)}\varepsilon +  \Delta (n_0, \delta), \forall m \geq n_0 \bigg)\\
    & \geq 1 - P \bigg(  \Vert \theta_{n_0} - \theta^{*} \Vert > \varepsilon \bigg)  - P\bigg( \theta_m^{'} >  \Delta (n_0, \delta)    \bigg)\\
    & \geq 1 - P \bigg(  \Vert \theta_{n_0} - \theta^{*} \Vert > \varepsilon \bigg)  -\sum\limits_{k = n_0 }^{ m-1} p_{k}    \\
    & \geq  1 - P \bigg(  \Vert \theta_{n_0} - \theta^{*} \Vert > \varepsilon \bigg)   -8\sum\limits_{k = n_0 }^{ m-1} e^{-\frac{D\delta^4}{ \omega(k)}}\\
    &\geq 1 - P \bigg(  \Vert \theta_{n_0} - \theta^{*} \Vert > \varepsilon \bigg)   -8\sum\limits_{k \geq  n_0 } e^{-\frac{D\delta^4}{ \omega(k)}}\\
\end{align*}

We have used the below lemma in the last inequality.
\begin{lemma}\label{martingale_ineq}

There exists positive constant $D$ such that for $ 0 < \delta \leq 1 $
    \begin{align*}
        p_{n-1} \leq 8 e^{-\frac{D\delta^4}{ \omega(n-1)}} 
    \end{align*}

where , 
\[
\omega(m)=
\begin{cases}
(n_0)^{c_\beta - \sigma}m^{-c_\beta}, & \text{if } c_\beta \leq \sigma, \\
 m ^{-\sigma},  & \text{if } otherwise.
\end{cases}
\]
\end{lemma}

\subsection{Proof of Corollary}

We use $\varsigma_{i}, i = 1,2,....$ to denote different constants in the proof.
For $\sigma = 1$ and $c_\beta$ sufficiently large, we have $\omega(m) = 1/m.$ Hence,
\begin{align*}
    8\sum\limits_{m \geq n_0} e^{-\frac{D\delta^4}{ \omega(m)}} = 8\sum\limits_{m \geq n_0} e^{-D m \delta^4  } \leq \varsigma_1 e^{-D n_0 \delta^4  }
\end{align*}

Let $\epsilon_1 = \varsigma_1 e^{-D n_0 \delta^4  } $ which gives us $\delta = \varsigma_2 n_0^{-1/4}\log ^{1/4} \bigg( \frac{\varsigma_1}{\epsilon_1}\bigg) $. This choice of $\delta$ gives us 
\begin{align*}
     8\sum\limits_{m \geq n_0} e^{-\frac{D\delta^4}{ \omega(m)}} \leq \epsilon_1
\end{align*}

We assume $c_\beta ( 1- \alpha) > 1$.

Now, 
\begin{align*}
    &\exp\bigg(-(1-\alpha)b_{n_0}(m-1)\bigg)\\
    =&\exp \bigg( -(1-\alpha)\sum_{k = n_0}^{k = m-1} \beta_k \bigg)\\
    \leq & \bigg(\frac{n_0 + 1}{m+1}\bigg)^{c_\beta(1-\alpha)}\\
    \leq & \frac{n_0 + 1}{m+1}
\end{align*}

Hence we have , $\forall m \geq n_0 > 0$ and $\beta_{n_0} < 1$,

\begin{align*}
    &P\bigg( \Vert \theta_m - \theta^{*} \Vert = O \bigg( \frac{n_0}{m} \varepsilon + \bigg(\frac{1}{n_0}\bigg)^{1/4} \log^{1/4} \bigg( \frac{1}{\epsilon_1   }\bigg)  + n_0^{(\nu - 1)/2
    } + n_0^{-\nu/2} + \sqrt{\epsilon_{\text{app}}} + \sqrt{bk}\bigg) \bigg)\\
    &\geq 1 - \epsilon_1- P \bigg(  \Vert \theta_{n_0} - \theta^{*} \Vert > \varepsilon \bigg)
\end{align*}

From assumption \ref{ergodicity} we have $b \leq \epsilon_{\text{app}}$. Hence we have, $\forall m \geq n_0 > 0,$
\begin{align*}
    & P\bigg( \Vert \theta_m - \theta^{*} \Vert = O \bigg( \frac{n_0}{m} \varepsilon + \bigg(\frac{1}{n_0}\bigg)^{1/4} \log^{1/4} \bigg( \frac{1}{\epsilon_1   }\bigg)  + n_0^{(\nu - 1)/2
    } + n_0^{-\nu/2} +  \sqrt{\epsilon_\text{app}}\bigg) \bigg)\\
    &\qquad \geq 1 - \epsilon_1-  P \bigg(  \Vert \theta_{n_0} - \theta^{*} \Vert > \varepsilon \bigg)
\end{align*}

Now , for $\varepsilon \geq 1$ , we have $\varepsilon \leq \varepsilon^2$. Hence,
\begin{align*}
    & \Vert \theta_{n_0} - \theta^{*} \Vert^2 > \epsilon^2 \implies  \Vert \theta_{n_0} - \theta^{*} \Vert^2 > \epsilon\\
\end{align*}
 which implies the below:
 \begin{align*}
     & \bigg\{ \Vert \theta_{n_0} - \theta^{*} \Vert^2 > \epsilon^2 \bigg\} \subseteq \bigg\{\Vert \theta_{n_0} - \theta^{*} \Vert^2 > \epsilon  \bigg\}\\
     \Rightarrow & P \bigg(\Vert \theta_{n_0} - \theta^{*} \Vert^2 > \epsilon^2  \bigg) \leq P\bigg(\Vert \theta_{n_0} - \theta^{*} \Vert^2 > \epsilon  \bigg)\\
 \end{align*}

Now , let $\varepsilon = \frac{\EE[\Vert \theta_{n_0} - \theta^{*}\Vert^2]}{\epsilon_2}$. Then we have ,
\begin{align*}
    P \bigg(  \Vert \theta_{n_0} - \theta^{*} \Vert > \varepsilon \bigg)
    =& P \bigg(  \Vert \theta_{n_0} - \theta^{*} \Vert^2 > \varepsilon^2 \bigg)\\
    \leq & P \bigg(  \Vert \theta_{n_0} - \theta^{*} \Vert^2 > \varepsilon \bigg)\\
    =&  P \bigg(  \Vert \theta_{n_0} - \theta^{*} \Vert^2 > \frac{\EE[\Vert \theta_{n_0} - \theta^{*}\Vert^2]}{\epsilon_2}\bigg) \leq \epsilon_2
\end{align*}

So we have , $\forall k \geq n_0 > 0,$

\begin{align*}
    &P\bigg( \Vert \theta_k - \theta^{*} \Vert = O \bigg( \frac{n_0}{k} \frac{\EE[\Vert \theta_{n_0} - \theta^{*}\Vert^2]}{\epsilon_2}+ \bigg(\frac{1}{n_0}\bigg)^{1/4} \log^{1/4} \bigg( \frac{1}{\epsilon_1  }\bigg)  + n_0^{(\nu - 1)/2
    } + n_0^{-\nu/2} \\
    &\qquad +  \sqrt{\epsilon_\text{app}}\ \bigg) \bigg) \geq 1 - \epsilon_1 - \epsilon_2 
\end{align*}

Now, we use the result of the following lemma.

\begin{lemma}\label{theta_bound}
For all $n \geq 1$ such that $2m\beta_{n} < 1$ and $\sigma = 1$, we have,
    \begin{align*}
        \EE[\Vert  \theta_{n} - \theta^{*} \Vert^2] =  \mathcal{O}(n^{-\nu/2}) + \mathcal{O}(n^{(\nu-1)/2})  .
    \end{align*}
\end{lemma}

Optimising over the values of $\nu$, we have $\nu = 0.5$.

Hence finally we have, (ignoring the approximation error)
$\forall k \geq n_0,$ and sufficiently large $n_0$ such that $2m\beta_{n_0} < 1,$

\begin{align*}
    &P\bigg( \Vert \theta_k - \theta^{*} \Vert = O \bigg( \frac{n_0^{3/4}}{k} \frac{1}{\sqrt{\epsilon_2}}+   \bigg(\frac{1}{n_0}\bigg)^{1/4} \log^{1/4} \bigg( \frac{1}{\epsilon_1  }\bigg)   + \bigg(\frac{1}{n_0} \bigg)^{1/4} \bigg) \bigg)\\
    &\qquad \geq 1 - \epsilon_1 - \epsilon_2 
\end{align*}

\subsection{Proof of Lemmas}

\subsubsection{Proof of Lemma \ref{lipschitz_L}}

We have,

\begin{align*}
    \nabla L (\theta) &= \sum_{s}\sum_{a}\mu_{\theta}(s)\pi_{\theta}(a|s)(r(s,a)  - L(\theta) + \sum_{s^{'}}P(s^{'}|s,a)V^{\theta}(s^{'}) - V^{\theta}(s)) \nabla \log \pi_{\theta}(a|s)\\
    &= \sum_{s}\sum_{a}\mu_{\theta}(s)\pi_{\theta}(a|s)(r(s,a)   + \sum_{s^{'}}P(s^{'}|s,a)V^{\theta}(s^{'})) \nabla \log \pi_{\theta}(a|s)\\
    &\qquad - \sum_{s}\sum_{a}\mu_{\theta}(s)\pi_{\theta}(a|s)( L(\theta) +  V^{\theta}(s)) \nabla \log \pi_{\theta}(a|s)\\
    &= \sum_{s}\sum_{a}\mu_{\theta}(s)\pi_{\theta}(a|s)(r(s,a)   + \sum_{s^{'}}P(s^{'}|s,a)V^{\theta}(s^{'})) \nabla \log \pi_{\theta}(a|s)\\
    &\qquad - \sum_{s}\mu_{\theta}(s)( L(\theta) +  V^{\theta}(s)) \sum_{a} \pi_{\theta}(a|s)\nabla \log \pi_{\theta}(a|s)\\
    &= \sum_{s}\sum_{a}\mu_{\theta}(s)\pi_{\theta}(a|s)(r(s,a)   + \sum_{s^{'}}P(s^{'}|s,a)V^{\theta}(s^{'})) \nabla \log \pi_{\theta}(a|s)\\
    &\qquad - \sum_{s}\mu_{\theta}(s)( L(\theta) +  V^{\theta}(s)) \sum_{a} \nabla  \pi_{\theta}(a|s)\\
    &= \sum_{s}\sum_{a}\mu_{\theta}(s)\pi_{\theta}(a|s)(r(s,a)   + \sum_{s^{'}}P(s^{'}|s,a)V^{\theta}(s^{'})) \nabla \log \pi_{\theta}(a|s)\\
    &\qquad - \sum_{s}\mu_{\theta}(s)( L(\theta) +  V^{\theta}(s)) \times 0\\
     &= \sum_{s}\sum_{a}\mu_{\theta}(s)\pi_{\theta}(a|s)(r(s,a)   + \sum_{s^{'}}P(s^{'}|s,a)V^{\theta}(s^{'})) \nabla \log \pi_{\theta}(a|s)\\
     &= \sum_{s}\sum_{a}\mu_{\theta}(s)(r(s,a)   + \sum_{s^{'}}P(s^{'}|s,a)V^{\theta}(s^{'})) \nabla  \pi_{\theta}(a|s)
\end{align*}

Now,
\begin{align*}
    &\Vert \nabla L(\theta_1) - L(\theta_2)\Vert\\
    =& \bigg\Vert \sum_{s}\sum_{a}\mu_{\theta_1}(s)\bigg(r(s,a)   + \sum_{s^{'}}P(s^{'}|s,a)V^{\theta_1}(s^{'})\bigg) \nabla  \pi_{\theta_1}(a|s)\\
    &\qquad - \sum_{s}\sum_{a}\mu_{\theta_2}(s)\bigg(r(s,a)   + \sum_{s^{'}}P(s^{'}|s,a)V^{\theta_2}(s^{'})\bigg) \nabla  \pi_{\theta_2}(a|s)   \bigg\Vert
\end{align*}

Let $X_{\theta}(s,a) = r(s,a)   + \sum_{s^{'}}P(s^{'}|s,a)V^{\theta}(s^{'})$.

Hence we have ,
\begin{align*}
   &\Vert \nabla L(\theta_1) - L(\theta_2)\Vert\\
   =&  \bigg\Vert \sum_{s}\sum_{a}\mu_{\theta_1}(s)X_{\theta_1}(s,a) \nabla  \pi_{\theta_1}(a|s)- \sum_{s}\sum_{a}\mu_{\theta_2}(s) X_{\theta_2}(s,a) \nabla  \pi_{\theta_2}(a|s)   \bigg\Vert\\
   =& \bigg\Vert \sum_{s}\sum_{a}\mu_{\theta_1}(s)X_{\theta_1}(s,a) \nabla  \pi_{\theta_1}(a|s)
 - \sum_{s}\sum_{a}\mu_{\theta_2}(s)X_{\theta_1}(s,a) \nabla  \pi_{\theta_1}(a|s)\\
   &\qquad + \sum_{s}\sum_{a}\mu_{\theta_2}(s)X_{\theta_1}(s,a) \nabla  \pi_{\theta_1}(a|s) - \sum_{s}\sum_{a}\mu_{\theta_2}(s) X_{\theta_2}(s,a) \nabla  \pi_{\theta_2}(a|s)   \bigg\Vert\\
    =& \bigg\Vert \sum_{s}\sum_{a}\mu_{\theta_1}(s)X_{\theta_1}(s,a) \nabla  \pi_{\theta_1}(a|s)
 - \sum_{s}\sum_{a}\mu_{\theta_2}(s)X_{\theta_1}(s,a) \nabla  \pi_{\theta_1}(a|s)\\
   &\qquad + \sum_{s}\sum_{a}\mu_{\theta_2}(s)X_{\theta_1}(s,a) \nabla  \pi_{\theta_1}(a|s) - \sum_{s}\sum_{a}\mu_{\theta_2}(s) X_{\theta_2}(s,a) \nabla  \pi_{\theta_1}(a|s)   \\
   &\qquad + \sum_{s}\sum_{a}\mu_{\theta_2}(s) X_{\theta_2}(s,a) \nabla  \pi_{\theta_1}(a|s) - \sum_{s}\sum_{a}\mu_{\theta_2}(s) X_{\theta_2}(s,a) \nabla  \pi_{\theta_2}(a|s) \bigg\Vert \\
   \leq & \bigg\Vert \sum_{s}\sum_{a}\mu_{\theta_1}(s)X_{\theta_1}(s,a) \nabla  \pi_{\theta_1}(a|s)
 - \sum_{s}\sum_{a}\mu_{\theta_2}(s)X_{\theta_1}(s,a) \nabla  \pi_{\theta_1}(a|s)\bigg\Vert\\
   &\qquad + \bigg\Vert\sum_{s}\sum_{a}\mu_{\theta_2}(s)X_{\theta_1}(s,a) \nabla  \pi_{\theta_1}(a|s) - \sum_{s}\sum_{a}\mu_{\theta_2}(s) X_{\theta_2}(s,a) \nabla  \pi_{\theta_1}(a|s)  \bigg\Vert \\
   &\qquad + \bigg\Vert\sum_{s}\sum_{a}\mu_{\theta_2}(s) X_{\theta_2}(s,a) \nabla  \pi_{\theta_1}(a|s) - \sum_{s}\sum_{a}\mu_{\theta_2}(s) X_{\theta_2}(s,a) \nabla  \pi_{\theta_2}(a|s) \bigg\Vert \\
   =& \bigg\Vert \sum_{s}\sum_{a}(\mu_{\theta_1}(s) - \mu_{\theta_2}(s))\pi_{\theta_1}(a|s)X_{\theta_1}(s,a) \nabla \log  \pi_{\theta_1}(a|s)\bigg\Vert\\
   &\qquad + \bigg\Vert\sum_{s}\sum_{a}\mu_{\theta_2}(s)(X_{\theta_1}(s,a) - X_{\theta_2}(s,a)) \nabla  \pi_{\theta_1}(a|s) \bigg\Vert \\
   &\qquad + \bigg\Vert\sum_{s}\sum_{a}\mu_{\theta_2}(s) X_{\theta_2}(s,a) (\nabla  \pi_{\theta_1}(a|s) - \nabla  \pi_{\theta_2}(a|s))  \bigg\Vert \\
   =& \underbrace{\bigg\Vert \sum_{s}(\mu_{\theta_1}(s) - \mu_{\theta_2}(s))\sum_{a}\pi_{\theta_1}(a|s)X_{\theta_1}(s,a) \nabla \log  \pi_{\theta_1}(a|s)\bigg\Vert}_{I_a}\\
   &\qquad + \underbrace{\bigg\Vert\sum_{s}\sum_{a}\mu_{\theta_2}(s)(X_{\theta_1}(s,a) - X_{\theta_2}(s,a)) \nabla  \pi_{\theta_1}(a|s) \bigg\Vert}_{I_b} \\
   &\qquad + \underbrace{\bigg\Vert\sum_{s}\sum_{a}\mu_{\theta_2}(s) X_{\theta_2}(s,a) (\nabla  \pi_{\theta_1}(a|s) - \nabla  \pi_{\theta_2}(a|s))  \bigg\Vert}_{I_c} \\
\end{align*}

For term $I_a$, we have ,
\begin{align*}
    &\bigg\Vert \sum_{s}(\mu_{\theta_1}(s) - \mu_{\theta_2}(s))\sum_{a}\pi_{\theta_1}(a|s)X_{\theta_1}(s,a) \nabla \log  \pi_{\theta_1}(a|s)\bigg\Vert\\
    \leq & \bigg\Vert \sum_{s}(\mu_{\theta_1}(s) - \mu_{\theta_2}(s))\sum_{a}\pi_{\theta_1}(a|s)X_{\theta_1}(s,a) \nabla \log  \pi_{\theta_1}(a|s)\bigg\Vert\\
    \leq & BU_{X}\sum_{s}\vert\mu_{\theta_1}(s) - \mu_{\theta_2}(s)\vert\\
    \leq & L_{\mu}BU_{X}\Vert \theta_1 - \theta_2\Vert
\end{align*}

For term $I_b$ we have ,
\begin{align*}
    &\bigg\Vert\sum_{s}\sum_{a}\mu_{\theta_2}(s)(X_{\theta_1}(s,a) - X_{\theta_2}(s,a)) \pi_{\theta_1}(s|a)\nabla \log \pi_{\theta_1}(a|s) \bigg\Vert\\
    \leq & L_{v}B\Vert \theta_1 - \theta_2 \Vert
\end{align*}

For term $I_c$ we have ,
\begin{align*}
    &\bigg\Vert\sum_{s}\sum_{a}\mu_{\theta_2}(s) X_{\theta_2}(s,a) (\nabla  \pi_{\theta_1}(a|s) - \nabla  \pi_{\theta_2}(a|s))  \bigg\Vert\\
    =& \bigg\Vert\sum_{s}\sum_{a}\mu_{\theta_2}(s) X_{\theta_2}(s,a) (\pi_{\theta_1}(a|s)\nabla  \log \pi_{\theta_1}(a|s) - \pi_{\theta_2}(a|s)\nabla \log  \pi_{\theta_2}(a|s))  \bigg\Vert\\
    \leq & \bigg\Vert\sum_{s}\sum_{a}\mu_{\theta_2}(s) X_{\theta_2}(s,a) (\pi_{\theta_1}(a|s)\nabla  \log \pi_{\theta_1}(a|s) - \pi_{\theta_2}(a|s)\nabla \log  \pi_{\theta_1}(a|s))  \bigg\Vert\\
    &\qquad + \bigg\Vert\sum_{s}\sum_{a}\mu_{\theta_2}(s) X_{\theta_2}(s,a) (\pi_{\theta_2}(a|s)\nabla  \log \pi_{\theta_1}(a|s) - \pi_{\theta_2}(a|s)\nabla \log  \pi_{\theta_2}(a|s))  \bigg\Vert\\
    \leq & BU_{X}L_{\pi}\Vert \theta_1 - \theta_2 \Vert + U_{X}K \Vert \theta_1 - \theta_2 \Vert
\end{align*}

Hence we have ,
\begin{align*}
    \Vert \nabla L(\theta_1) - L(\theta_2)\Vert \leq (L_{\mu}BU_{X}+ L_{v}B + BU_{X}L_{\pi} +U_{X}K )\Vert \theta_1 - \theta_2 \Vert
\end{align*}

Now , we have ,
\begin{align*}
     V^{\theta}(s) &= \EE\bigg[\sum\limits_{t=0}^{\infty}(r(s_t,a_t) - L(\theta)) |s_0 = s \bigg]\\
     &= \sum\limits_{t=0}^{\infty}\EE[r(s_t,a_t) - L(\theta) | s_0 = s ]\\
     &\leq \frac{2U_{r}b}{1 - k}
\end{align*}

Hence we have $U_{X} = U_{r} + \frac{2U_{r}b}{1 - k}. $

So finally we have,
\begin{align*}
    \Vert \nabla L(\theta_1) - L(\theta_2)\Vert \leq M_{L}\Vert \theta_1 - \theta_2 \Vert
\end{align*}
where 
\begin{align*}
    M_{L} &= L_{\mu}BU_{X}+ L_{v}B + BU_{X}L_{\pi} +U_{X}K\\
    &= U_{X}(L_{\mu}B + BL_{\pi} + K) + L_{v}B\\
    &= U_{r}\bigg( 1 + \frac{2b}{1-k} \bigg)(L_{\mu}B + BL_{\pi} + K) + L_{v}B
\end{align*}

\subsubsection{Proof of lemma \ref{lipschitz_mu_pi_V}}

The  first inequality follows from lemma B.1 of \cite{wu2022finite}. The proof for second inequality is as follows :

For finite state and action MDPs, under bounded costs and with  policy continuously differentiable and which selects each action with a positive probability, one can show that the stationary distribution and so the average reward is continuously differentiable and bounded, making it Lipschitz (see Theorem A.1, \cite{panda_bhatnagar_2025_CA}) . The differential value function can be seen to then be differentiable from the fact that 
\[
\sum_{k=0}^K P_\pi^k(H^\pi - Q(\pi)) = V^\pi-P_\pi^{K+1}V^\pi,
\]
where $H^\pi = E[r |X_0=i, \pi]$ and $Q(\pi) = J(\pi)$, 
see p.284,\cite{Bertsekas2013OptimalControlVol2}, upon noting that $\lim_{K\rightarrow\infty} P_\pi^{K+1}V^\pi=0$.\\
The third inequality follows from assumption \ref{assum:policy-lipschitz-bounded}(c)
 
\subsubsection{Proof of Lemma \ref{average_reward_convergence}} \label{proof_avg_reward}

Let us denote $y_n = L_n - L(\theta_n)$. 

So we have,
\begin{align*}
    &y_{n+1} = y_n + L(\theta_{n}) - L(\theta_{n+1}) + \gamma_n(r_n - L_n)\\
    \Rightarrow & y_{n+1}^2 = ( y_n + L(\theta_{n}) - L(\theta_{n+1}) + \gamma_n(r_n - L_n))^2\\
    \Rightarrow &  y_{n+1}^2 \leq  y_n^2+2y_n(L(\theta_n)-L(\theta_{n+1}))+2\gamma_n y_n(r_n-L_n)+2(L(\theta_n)-L(\theta_{n+1}))^2+2\gamma_n^2(r_n-L_n)^2\\
    \Rightarrow & y_{t+1}^2 \leq (1-2\gamma_t)y_t^2+2\gamma_ty_t(r_t-L(\theta_t))+2y_t(L(\theta_t)-L(\theta_{t+1}))+2(L(\theta_t)-L(\theta_{t+1}))^2\\
    &\qquad+2\gamma_t^2(r_t-L_t)^2.
\end{align*}

where $r_n = r(s_n,a_n)$.

After rearranging the terms we have,

\begin{align*}
    y_m^2 &\leq \frac{y_m^2 - y_{m+1}^2}{2\gamma_m} +  y_m(r_m-L(\theta_m))+ \frac{y_m}{\gamma_m}(L(\theta_m)-L(\theta_{m+1}))+\frac{1}{\gamma_m}(L(\theta_m)-L(\theta_{m+1}))^2\\
    &\qquad+\gamma_m(r_m-L_m)^2\\  
\end{align*}

Now we have,
\begin{align}
    \sum_{k = n_0}^{k = m}\xi(m,k+1)\beta_k y_k^2 &\leq \underbrace{\sum_{k = n_0}^{k = m}\xi(m,k+1)\beta_k\frac{y_k^2 - y_{k+1}^2}{2\gamma_k}}_{I_a} +  \underbrace{\sum_{k = n_0}^{k = m}\xi(m,k+1)\beta_k y_k(r_k-L(\theta_k))}_{I_b} \notag\\
    &\qquad + \underbrace{\sum_{k = n_0}^{k = m}\xi(m,k+1)\beta_k\frac{y_k}{\gamma_k}(L(\theta_k)-L(\theta_{k+1}))}_{I_c}\notag\\
    &\qquad + \underbrace{\sum_{k = n_0}^{k = m}\xi(m,k+1)\beta_k\frac{1}{\gamma_k}(L(\theta_k)-L(\theta_{k+1}))^2}_{I_d} \notag\\ 
    &\qquad+  \underbrace{\sum_{k = n_0}^{k = m}\xi(m,k+1)\beta_k\gamma_k(r_k-L_k)^2 }_{I_e}\label{ineq_I_3}
\end{align}

For term $I_a$ we have,


\begin{align*}
    I_a &= \sum_{k = n_0}^{k = m}\xi(m,k+1)\beta_k\frac{y_k^2 - y_{k+1}^2}{2\gamma_k}\\
    &= \sum_{k = n_0}^{k = m}\xi(m,k+1)\frac{y_k^2 - y_{k+1}^2}{2\frac{\gamma_k}{\beta_k}}\\
    &= \sum_{k = n_0}^{k = m}\frac{\xi(m,k+1) y_k^2 - \xi(m,k+1) y_{k+1}^2}{2\frac{\gamma_k}{\beta_k}}\\
    &= \sum_{k = n_0}^{k = m}\frac{\xi(m,k) y_k^2 - \xi(m,k+1) y_{k+1}^2}{2\frac{\gamma_k}{\beta_k}} + \sum_{k = n_0}^{k = m}\frac{(\xi(m,k + 1) -\xi(m,k) ) y_k^2 }{2\frac{\gamma_k}{\beta_k}}\\
    &=  \underbrace{\sum_{k = n_0}^{k = m}\frac{\xi(m,k) y_k^2 - \xi(m,k+1) y_{k+1}^2}{2\frac{\gamma_k}{\beta_k}}}_{I_{a1}} + \underbrace{\mathcal{O}\bigg( \sum_{k = n_0}^{k = m} \xi(m,k+1)\beta_k \frac{\beta_k}{\gamma_k} \bigg)}_{I_{a2}}\\
    &= \mathcal{O}\bigg(\frac{\beta_{n_0 - 1}}{\gamma_{n_0 - 1}}\bigg) + \mathcal{O}(n_0^{\nu - \sigma})
\end{align*}

Term $I_{a1}$ can be analysed similar to term $I_1$ in page number 22 of \cite{wu2022finite}. Term $I_{a2}$ can be analysed similar to term $I_2$  in page number 22 of \cite{wu2022finite}.

For term $I_b$ we have,
\begin{align*}
    &\sum_{k = n_0}^{k = m}\xi(m,k+1)\beta_k y_k(r(s_k,a_k)-L(\theta_k))\\
    &= \sum_{k = n_0}^{k = m}\xi(m,k+1)\beta_k y_k( r(s_k,a_k) - \sum_{a}\pi_{\theta_k}(a|s_k)r(s_k,a))\\
    &\qquad + \sum_{k = n_0}^{k = m}\xi(m,k+1)\beta_k y_k( \sum_{a}\pi_{\theta_k}(a|s_k)r(s_k,a) - \sum_{s , a }\mu_{\theta_{k}}(s)\pi_{\theta_{k}}(a|s)r(s,a) )\\
    &= \sum_{k = n_0}^{k = m}\xi(m,k+1)\beta_k y_k( r(s_k,a_k) - \sum_{a}\pi_{\theta_k}(a|s_k)r(s_k,a))\\
    &\qquad + \sum_{k = n_0}^{k = m}\xi(m,k+1)\beta_k y_k( \bar{r}_{\theta_k}(s_k) - \sum_{s}\mu_{\theta_k}(s) \bar{r}_{\theta_k}(s))\\
    &\leq  \sum_{k = n_0}^{k = m}\xi(m,k+1)\beta_k y_k \bigg( r(s_k,a_k) - \sum_{a}\pi_{\theta_k}(a|s_k)r(s_k,a) + \bar{r}_{\theta_k}(s_{k+1}) -\sum_{s }P^{\theta_k}(s |s_k)\bar{r}_{\theta_k}(s)\bigg)\\
    &\qquad +  \mathcal{O}(\beta_{n_0}) + \mathcal{O}(n_0^{-\nu }) + \mathcal{O}(bk)
\end{align*}

where we define $\bar{r}_{\theta}(s) =  \sum_{a}\pi_{\theta}(a|s)r(s,a) .$

Term $I_b$ above can be analysed similar to term $I_1$ in (\ref{ineq_second_bound}).

For term $I_c$ we have,
\begin{align*}
    &\sum_{k = n_0}^{k = m}\xi(m,k+1)\beta_k\frac{y_k}{\gamma_k}(L(\theta_k)-L(\theta_{k+1}))\\
    =& \mathcal{O}\bigg( \sum_{k = n_0}^{k = m}\frac{\xi(m,k+1) \beta_k^2}{\gamma_k}\bigg)\\
    =& \mathcal{O}(n_0^{\nu - \sigma})
\end{align*}

For term $I_d$ we have,
\begin{align*}
    &\sum_{k = n_0}^{k = m}\xi(m,k+1)\beta_k\frac{1}{\gamma_k}(L(\theta_k)-L(\theta_{k+1}))^2\\
    &= \mathcal{O}\bigg( \sum_{k = n_0}^{k = m}\xi(m,k+1)\beta_k \frac{\beta_k^2}{\gamma_k} \bigg)\\
    &= \mathcal{O}(n_0^{\nu - 2\sigma}).\\
\end{align*}

For term $I_e$ we have,
\begin{align*}
   \sum_{k = n_0}^{k = m}\xi(m,k+1)\beta_k\gamma_k(r_k-L_k)^2 = \mathcal{O}\bigg(\sum_{k = n_0}^{k = m} \xi(m,k+1)\beta_k \gamma_k\bigg) = \mathcal{O}(n_0^{- \nu })\\
\end{align*}

Hence after gathering all the terms we have ,

\begin{align*}
    &\sum_{k = n_0}^{k = m}\xi(m,k+1)\beta_k y_k^2\\
    & \leq  \mathcal{O}(n_0^{\nu - \sigma})  + \mathcal{O}(n_0^{-\nu }) + \mathcal{O}(bk)\\
    &\qquad + \sum_{k = n_0}^{k = m}\xi(m,k+1)\beta_k y_k \bigg( r(s_k,a_k) - \sum_{a}\pi_{\theta_k}(a|s_k)r(s_k,a) + \bar{r}_{\theta_k}(s_{k+1}) -\sum_{s }P^{\theta_k}(s |s_k)\bar{r}_{\theta_k}(s)\bigg)
\end{align*}

\subsubsection{Proof of Lemma \ref{critic_convergence}}

We have the following update rule for critic :
\begin{align*}
    v_{t+1} = \Gamma(v_t + \alpha_t \delta_t \phi(s_t))
\end{align*}

Let $z_t = v_t - v^{*}(\theta_t)$. So we have , 
\begin{align*}
    \Vert z_{t+1}\Vert^2 &= \Vert v_{t+1} - v^{*}(\theta_{t+1})\Vert^2\\
    &= \Vert \Gamma(v_t + \alpha_t \delta_t \phi(s_t)) - v^{*}(\theta_{t+1})\Vert^2\\
    &\leq \Vert v_t + \alpha_t \delta_t \phi(s_t) - v^{*}(\theta_{t+1})\Vert^2\\
    &= \Vert v_t - v^{*}(\theta_t) + \alpha_t \delta_t \phi(s_t) - v^{*}(\theta_{t+1}) +  v^{*}(\theta_{t})\Vert^2\\
    &= \Vert z_t + \alpha_t \delta_t \phi(s_t) - v^{*}(\theta_{t+1}) +  v^{*}(\theta_{t})\Vert^2\\
    &= \Vert z_t\Vert^2 + 2 \alpha_t\langle z_t , \delta_t \phi(s_t)  \rangle  + 2 \langle z_t ,v^{*}(\theta_{t}) - v^{*}(\theta_{t+1}) \rangle + \Vert  \alpha_t \delta_t \phi(s_t) + ( v^{*}(\theta_{t}) -  v^{*}(\theta_{t+1})) \Vert^2 \\
    &\leq  \Vert z_t\Vert^2 +  2 \alpha_t\langle z_t , \delta_t \phi(s_t) - E_{\theta_t}[\delta_t \phi(s_t)]  \rangle + 2 \alpha_t\langle z_t , E_{\theta_t}[\delta_t \phi(s_t)]  \rangle + 2 \langle z_t ,v^{*}(\theta_{t}) - v^{*}(\theta_{t+1}) \rangle\\
    &\qquad + 2\alpha_t^2 \Vert \delta_t \phi(s_t) \Vert^2 + 2 \Vert v^{*}(\theta_t) - v^{*}(\theta_{t+1})\Vert^2
\end{align*}

Now we have ,
\begin{align*}
   &\langle z_t , \EE_{\theta_t}[\delta_t \phi(s_t)]  \rangle \\
   =& \langle z_t , \EE_{\theta_t}[(r(s_t , a_t) - L_t + \phi(s_{t+1})^T v_t - \phi(s_{t})^T v_t) \phi(s_t)]  \rangle\\
   =&   \langle z_t , \EE_{\theta_t}[(r(s_t , a_t) - L(\theta_t) + \phi(s_{t+1})^T v_t - \phi(s_{t})^T v_t) \phi(s_t)]  \rangle + \langle z_t , (L(\theta_t) - L_t)\EE_{\theta_t}[\phi(s_t)] \rangle\\
   =&  \langle z_t , \EE_{\theta_t}[(r(s_t , a_t) - L(\theta_t) + \phi(s_{t+1})^T v_t - \phi(s_{t})^T v_t) \phi(s_t)]  \rangle\\
   &\qquad +  \langle z_t , \EE_{\theta_t}[(r(s_t , a_t) - L(\theta_t) + \phi(s_{t+1})^T v^{*}(\theta_t) - \phi(s_{t})^T v^{*}(\theta_t)) \phi(s_t)]  \rangle\\
   &\qquad + \langle z_t , (L(\theta_t) - L_t)E_{\theta_t}[\phi(s_t)] \rangle\\
   =& \langle z_t, \EE[(\phi(s_{t+1})^T -\phi(s_{t})^T ) (v_t - v^{*}(\theta_t)) \phi(s_t)] \rangle + \langle z_t , (L(\theta_t) - L_t)E_{\theta_t}[\phi(s_t)] \rangle\\
   \leq & -\lambda \Vert z_t \Vert^2 +  \langle z_t , (L(\theta_t) - L_t)E_{\theta_t}[\phi(s_t)] \rangle
\end{align*}

So finally we have ,
\begin{align*}
     \Vert z_{t+1}\Vert^2 &\leq  \Vert z_t\Vert^2 +  2 \alpha_t\langle z_t , \delta_t \phi(s_t) - E_{\theta_t}[\delta_t \phi(s_t)]  \rangle -2\alpha_t\lambda \Vert z_t \Vert^2 +  2\alpha_t\langle z_t , (L(\theta_t) - L_t)E_{\theta_t}[\phi(s_t)] \rangle \\
     &\qquad + 2 \langle z_t ,v^{*}(\theta_{t}) - v^{*}(\theta_{t+1}) \rangle
    + 2\alpha_t^2 \Vert \delta_t \phi(s_t) \Vert^2 + 2 \Vert v^{*}(\theta_t) - v^{*}(\theta_{t+1})\Vert^2
\end{align*}

After rearranging the terms we have ,

\begin{align*}
    2 \lambda \Vert z_t\Vert^2 &\leq \frac{ \Vert z_t\Vert^2 -  \Vert z_{t+1}\Vert^2}{\alpha_t} +  2 \langle z_t , \delta_t \phi(s_t) - E_{\theta_t}[\delta_t \phi(s_t)]  \rangle +  2\langle z_t , (L(\theta_t) - L_t)E_{\theta_t}[\phi(s_t)] \rangle \\
     &\qquad + \frac{2}{\alpha_t} \langle z_t ,v^{*}(\theta_{t}) - v^{*}(\theta_{t+1}) \rangle
    + 2\alpha_t \Vert \delta_t \phi(s_t) \Vert^2 + \frac{2}{\alpha_t} \Vert v^{*}(\theta_t) - v^{*}(\theta_{t+1})\Vert^2\\
    &\leq \frac{ \Vert z_t\Vert^2 -  \Vert z_{t+1}\Vert^2}{\alpha_t} + 2 \langle z_t , \delta_t \phi(s_t) - E_{\theta_t}[\delta_t \phi(s_t)]  \rangle + 2\Vert z_t \Vert \Vert L(\theta_t) - L_t \Vert\\
    &\qquad + C_1 \frac{\beta_t}{\alpha_t}\Vert z_t \Vert +  2\alpha_t \Vert \delta_t \phi(s_t) \Vert^2 + C_2 \frac{\beta_t^2}{\alpha_t}
\end{align*}

Hence we have ,

\begin{align*}
    \Rightarrow  2 \lambda \sum_{k = n_0}^{k = m}\xi(m,k+1)\beta_k \Vert z_k\Vert^2 &\leq \underbrace{\sum_{k = n_0}^{k = m}\xi(m,k+1)\beta_k \frac{ \Vert z_k\Vert^2 -  \Vert z_{k+1}\Vert^2}{\alpha_k}}_{I_1} \\
    &\qquad + \underbrace{2\sum_{k = n_0}^{k = m}\xi(m,k+1)\beta_k \langle z_k , \delta_k \phi(s_k) - E_{\theta_k}[\delta_k \phi(s_k)]  \rangle}_{I_2}\\
    &\qquad + \underbrace{2\sum_{k = n_0}^{k = m}\xi(m,k+1)\beta_k \Vert z_k \Vert \vert L(\theta_k) - L_k \vert}_{I_3}\\
    &\qquad + \underbrace{C_1\sum_{k = n_0}^{k = m}\xi(m,k+1)\beta_k  \frac{\beta_k}{\alpha_k}\Vert z_k \Vert }_{I_4}+ \underbrace{2\sum_{k = n_0}^{k = m}\xi(m,k+1)\beta_k  \alpha_k \Vert \delta_k \phi(s_k) \Vert^2}_{I_5}\\
    &\qquad + \underbrace{C_2\sum_{k = n_0}^{k = m}\xi(m,k+1)\beta_k \frac{\beta_k^2}{\alpha_k}}_{I_6}
\end{align*}

For term $I_1$ we have,
\begin{align*}
    &\sum_{k = n_0}^{k = m}\xi(m,k+1)\beta_k \frac{ \Vert z_k\Vert^2 -  \Vert z_{k+1}\Vert^2}{\alpha_k}\\
    &= \sum_{k = n_0}^{k = m}\xi(m,k+1) \frac{ \Vert z_k\Vert^2 -  \Vert z_{k+1}\Vert^2}{\frac{\alpha_k}{\beta_k}}\\
\end{align*}

Hence we have 
\begin{align*}
    I_1 = \mathcal{O}\bigg(\frac{\beta_{n_0 - 1}}{\alpha_{n_0 - 1}}\bigg) + \mathcal{O}(n_0^{\nu - \sigma})
\end{align*}

Term $I_1$ can be analysed similar to term $I_a$ in subsection \ref{proof_avg_reward}.

For term $I_2$ we have,
\begin{align*}
    I_2 = 2\sum_{k = n_0}^{k = m}\xi(m,k+1)\beta_k \langle z_k , \delta_k \phi(s_k) - E_{\theta_k}[\delta_k \phi(s_k)]  \rangle
\end{align*}

Let $W(s,a,s^{'}, L,v) = (r(s,a) - L + \phi(s^{'})^T v - \phi(s)^Tv)\phi(s)$.

Hence we have,
\begin{align*}
     &I_2\\
     &= 2\sum_{k = n_0}^{k = m}\xi(m,k+1)\beta_k \langle z_k , W(s_k,a_k,s_{k+1}, L_k,v_k) - \sum\limits_{s}\sum\limits_{a}\sum\limits_{s^{'}}\mu_{\theta_k}(s)\pi_{\theta_k}(a|s)P(s^{'}|s,a)W(s,a,s^{'}, L_k,v_k)  \rangle\\
     &= 2\sum_{k = n_0}^{k = m}\xi(m,k+1)\beta_k \langle z_k , W(s_k,a_k,s_{k+1}, L_k,v_k)  - \sum\limits_{a}\sum\limits_{s^{'}}\pi_{\theta_k}(a|s_k)P(s^{'}|s_k,a)W(s_k,a,s^{'}, L_k,v_k)\rangle \\
     &\qquad + 2\sum_{k = n_0}^{k = m}\xi(m,k+1)\beta_k \langle z_k ,\sum\limits_{a}\sum\limits_{s^{'}}\pi_{\theta_k}(a|s_k)P(s^{'}|s_k,a)W(s_k,a,s^{'}, L_k,v_k)\\
     &\qquad - \sum\limits_{s}\sum\limits_{a}\sum\limits_{s^{'}}\mu_{\theta_k}(s)\pi_{\theta_k}(a|s)P(s^{'}|s,a)W(s,a,s^{'}, L_k,v_k) \rangle \\
     &= 2\sum_{k = n_0}^{k = m}\xi(m,k+1)\beta_k \langle z_k , W(s_k,a_k,s_{k+1}, L_k,v_k)  - \sum\limits_{a}\sum\limits_{s^{'}}\pi_{\theta_k}(a|s_k)P(s^{'}|s_k,a)W(s_k,a,s^{'}, L_k,v_k)\rangle \\
     &\qquad + 2\sum_{k = n_0}^{k = m}\xi(m,k+1)\beta_k \langle z_k , \bar{W}_{\theta_k}(s_k,L_k , v_k) - \sum_{s} \mu_{\theta_k}(s) \bar{W}_{\theta_k}(s,L_k , v_k) \rangle \\
\end{align*}

where we define $\bar{W}_{\theta}(s,L,v) = \sum\limits_{a}\sum\limits_{s^{'}}\pi_{\theta}(a|s)P(s^{'}|s,a)W(s,a,s^{'}, L,v).$

After analysing term $I_2$ similar to term  $I_1$ in (\ref{ineq_second_bound}).
\begin{align*}
    &I_2 \\ 
    & \leq  2\sum_{k = n_0}^{k = m}\xi(m,k+1)\beta_k \langle z_k , W(s_k,a_k,s_{k+1}, L_k,v_k)  - \sum\limits_{a}\sum\limits_{s^{'}}\pi_{\theta_k}(a|s_k)P(s^{'}|s_k,a)W(s_k,a,s^{'}, L_k,v_k)\rangle \\
     &\qquad + 2\sum_{k = n_0}^{k = m}\xi(m,k+1)\beta_k \langle z_k , \bar{W}_{\theta_k}(s_{k+1},L_k , v_k) - \sum_{s} P^{\theta_k}(s|s_k) \bar{W}_{\theta_k}(s,L_k , v_k) \rangle \\
    &\qquad + 4\beta_{n_0 - 1}W_{max} + \mathcal{O}(n_0^{ - \nu}) + \mathcal{O}(bk)
\end{align*}

For term $I_3$ we have ,
\begin{align*}
     &2\sum_{k = n_0}^{k = m}\xi(m,k+1)\beta_k \vert L(\theta_k) - L_k \vert  \Vert z_k \Vert\\
    =& 2\sum_{k = n_0}^{k = m}\sqrt{\xi(m,k+1) \beta_k ( L(\theta_k) - L_k )^2}\sqrt{ \xi(m,k+1) \beta_k \Vert z_k \Vert^2}\\
    \leq & \sqrt{\sum_{k = n_0}^{k = m}\xi(m,k+1) \beta_k ( L(\theta_k) - L_k )^2}\sqrt{\sum_{k = n_0}^{k = m} \xi(m,k+1) \beta_k \Vert z_k \Vert^2}
\end{align*}

For term $I_4$ we have ,
\begin{align*}
    C_1\sum_{k = n_0}^{k = m}\xi(m,k+1)\beta_k  \frac{\beta_k}{\alpha_k}\Vert z_k \Vert = \mathcal{O}(n_0^{ \nu - \sigma})
\end{align*}

For term $I_5$ we have,
\begin{align*}
    2\sum_{k = n_0}^{k = m}\xi(m,k+1)\beta_k  \alpha_k \Vert \delta_k \phi(s_k) \Vert^2 = \mathcal{O}(n_0^{-\nu })
\end{align*}

For term $I_6$ we have,
\begin{align*}
    C_2\sum_{k = n_0}^{k = m}\xi(m,k+1)\beta_k \frac{\beta_k^2}{\alpha_k} = \mathcal{O}(n_0^{ \nu - 2\sigma})
\end{align*}

Hence after gathering all the terms we have,
\begin{align*}
    &\Rightarrow  2 \lambda \sum_{k = n_0}^{k = m}\xi(m,k+1)\beta_k \Vert z_k\Vert^2\\
    &\leq  \mathcal{O}(n_0^{\nu - \sigma}) \\
    &\qquad +  2\sum_{k = n_0}^{k = m}\xi(m,k+1)\beta_k \langle z_k , W(s_k,a_k,s_{k+1}, L_k,v_k)  - \sum\limits_{a}\sum\limits_{s^{'}}\pi_{\theta_k}(a|s_k)P(s^{'}|s_k,a)W(s_k,a,s^{'}, L_k,v_k)\rangle \\
     &\qquad + 2\sum_{k = n_0}^{k = m}\xi(m,k+1)\beta_k \langle z_k , \bar{W}_{\theta_k}(s_{k+1},L_k , v_k) - \sum_{s} P^{\theta_k}(s|s_k) \bar{W}_{\theta_k}(s,L_k , v_k) \rangle \\
    &\qquad +4\beta_{n_0 - 1}W_{max} + \mathcal{O}(n_0^{ - \nu}) + \mathcal{O}(bk)\\
    &\qquad+ \sqrt{\sum_{k = n_0}^{k = m}\xi(m,k+1) \beta_k ( L(\theta_k) - L_k )^2}\sqrt{\sum_{k = n_0}^{k = m}\xi(m,k+1) \beta_k \Vert z_k \Vert^2}  
\end{align*}

Let 
\begin{align*}
    A(m,n_0) &=  \mathcal{O}(n_0^{\nu - \sigma})\\
    &\qquad +  2\sum_{k = n_0}^{k = m}\xi(m,k+1)\beta_k \langle z_k , W(s_k,a_k,s_{k+1}, L_k,v_k)  - \sum\limits_{a}\sum\limits_{s^{'}}\pi_{\theta_k}(a|s_k)P(s^{'}|s_k,a)W(s_k,a,s^{'}, L_k,v_k)\rangle \\
     &\qquad + 2\sum_{k = n_0}^{k = m}\xi(m,k+1)\beta_k \langle z_k , \bar{W}_{\theta_k}(s_{k+1},L_k , v_k) - \sum_{s} P^{\theta_k}(s|s_k) \bar{W}_{\theta_k}(s,L_k , v_k) \rangle\\
    &\qquad + 4\beta_{n_0 - 1}W_{max} + \mathcal{O}(n_0^{ - \nu}) + \mathcal{O}(bk)
\end{align*}
and 

\begin{align*}
    B(m, n_0) = \sum_{k = n_0}^{k = m}\xi(m,k+1) \beta_k ( L(\theta_k) - L_k )^2
\end{align*}

Hence we have ,
\begin{align*}
    & 2 \lambda \sum_{k = n_0}^{k = m}\xi(m,k+1)\beta_k \Vert z_k\Vert^2 \leq A(m , n_0) +   \sqrt{B(m, n_0)}\sqrt{\sum_{k = n_0}^{k = m}\xi(m,k+1) \beta_k \Vert z_k \Vert^2} \\
    &\Rightarrow  \sum_{k = n_0}^{k = m}\xi(m,k+1)\beta_k \Vert z_k\Vert^2  - \frac{1}{2 \lambda}\sqrt{B(m, n_0)}\sqrt{\sum_{k = n_0}^{k = m}\xi(m,k+1) \beta_k \Vert z_k \Vert^2} \leq \frac{A(m , n_0)}{2 \lambda}\\
    &\Rightarrow \sum_{k = n_0}^{k = m}\xi(m,k+1)\beta_k \Vert z_k\Vert^2  - \frac{2}{4 \lambda}\sqrt{B(m, n_0)}\sqrt{\sum_{k = n_0}^{k = m}\xi(m,k+1) \beta_k \Vert z_k \Vert^2} + \frac{B(m,n_0)}{16 \lambda^2}\\
    &\qquad \leq \frac{A(m , n_0)}{2 \lambda} + \frac{B(m,n_0)}{16 \lambda^2}\\
    &\Rightarrow \bigg(\sqrt{\sum_{k = n_0}^{k = m}\xi(m,k+1)\beta_k \Vert z_k\Vert^2} - \frac{\sqrt{B(m,n_0)}}{4\lambda}\bigg)^2 \leq \frac{A(m , n_0)}{2 \lambda} + \frac{B(m,n_0)}{16 \lambda^2}\\
    &\Rightarrow \sqrt{\sum_{k = n_0}^{k = m}\xi(m,k+1)\beta_k \Vert z_k\Vert^2} - \frac{\sqrt{B(m,n_0)}}{4\lambda} \leq \sqrt{\frac{A(m , n_0)}{2 \lambda}} + \sqrt{\frac{B(m,n_0)}{16 \lambda^2}}\\
    &\Rightarrow \sqrt{\sum_{k = n_0}^{k = m}\xi(m,k+1)\beta_k \Vert z_k\Vert^2} \leq \sqrt{\frac{A(m , n_0)}{2 \lambda}} + 2\sqrt{\frac{B(m,n_0)}{16 \lambda^2}}\\
    &\Rightarrow \sum_{k = n_0}^{k = m}\xi(m,k+1)\beta_k \Vert z_k\Vert^2 \leq \frac{A(m , n_0)}{ \lambda} + \frac{1}{2\lambda}B(m,n_0)\\
    &\Rightarrow \sum_{k = n_0}^{k = m}\xi(m,k+1)\beta_k \Vert z_k\Vert^2
    \leq  \mathcal{O}(n_0^{\nu - \sigma})\\
    &\qquad +  \frac{2}{\lambda}\sum_{k = n_0}^{k = m}\xi(m,k+1)\beta_k \langle z_k , W(s_k,a_k,s_{k+1}, L_k,v_k)  - \sum\limits_{a}\sum\limits_{s^{'}}\pi_{\theta_k}(a|s_k)P(s^{'}|s_k,a)W(s_k,a,s^{'}, L_k,v_k)\rangle \\
     &\qquad + \frac{2}{\lambda}\sum_{k = n_0}^{k = m}\xi(m,k+1)\beta_k \langle z_k , \bar{W}_{\theta_k}(s_{k+1},L_k , v_k) - \sum_{s} P^{\theta_k}(s|s_k) \bar{W}_{\theta_k}(s,L_k , v_k) \rangle\\
    &\qquad +  \mathcal{O}(bk)  + \mathcal{O}(n_0^{-\nu}) + \frac{1}{2\lambda}\sum_{k = n_0}^{k = m}\xi(m,k+1) \beta_k ( L(\theta_k) - L_k )^2
\end{align*}
Now putting the results of lemma \ref{average_reward_convergence} in the above inequality we have ,

\begin{align*}
    &\sum_{k = n_0}^{k = m}\xi(m,k+1)\beta_k \Vert z_k\Vert^2 \leq \mathcal{O}(n_0^{\nu - \sigma})\\
    &\qquad +  \frac{2}{\lambda}\sum_{k = n_0}^{k = m}\xi(m,k+1)\beta_k \langle z_k , W(s_k,a_k,s_{k+1}, L_k,v_k)  - \sum\limits_{a}\sum\limits_{s^{'}}\pi_{\theta_k}(a|s_k)P(s^{'}|s_k,a)W(s_k,a,s^{'}, L_k,v_k)\rangle \\
     &\qquad + \frac{2}{\lambda}\sum_{k = n_0}^{k = m}\xi(m,k+1)\beta_k \langle z_k , \bar{W}_{\theta_k}(s_{k+1},L_k , v_k) - \sum_{s} P^{\theta_k}(s|s_k) \bar{W}_{\theta_k}(s,L_k , v_k) \rangle\\
    &\qquad + \mathcal{O}(bk)  + \mathcal{O}(n_0^{-\nu}) \\
    &\qquad + \frac{1}{2 \lambda}\sum_{k = n_0}^{k = m}\xi(m,k+1)\beta_k y_k \bigg( r(s_k,a_k) - \sum_{a}\pi_{\theta_k}(a|s_k)r(s_k,a) + \bar{r}_{\theta_k}(s_{k+1}) -\sum_{s }P^{\theta_k}(s |s_k)\bar{r}_{\theta_k}(s)\bigg) 
\end{align*}

\subsubsection{Proof of lemma \ref{contraction_F}}

We have ,

\begin{align*}
     &\sum_{s}\sum_{a}\mu_{\theta}(s)\pi_{\theta}(a|s)F(\theta,s,a)\\
     =& \sum_{s}\sum_{a}\mu_{\theta}(s)\pi_{\theta}(a|s)((r(s,a)  - L(\theta) + \sum_{s^{'}}P(s^{'}|s,a)V^{\theta}(s^{'}) - V^{\theta}(s)) \nabla \log \pi_{\theta}(a|s) + \theta)\\
     =& \sum_{s}\sum_{a}\mu_{\theta}(s)\pi_{\theta}(a|s)(r(s,a)  - L(\theta) + \sum_{s^{'}}P(s^{'}|s,a)V^{\theta}(s^{'}) - V^{\theta}(s)) \nabla \log \pi_{\theta}(a|s) + \theta\\
     =& \nabla L(\theta) + \theta
\end{align*}

Now ,

\begin{align*}
    &\bigg\Vert \sum_{s}\sum_{a}\mu_{\theta_1}(s)\pi_{\theta_1}(a|s)F(\theta_1,s,a) - \sum_{s}\sum_{a}\mu_{\theta_2}(s)\pi_{\theta_2}(a|s)F(\theta_2,s,a)\bigg\Vert^2\\
    =& \bigg\Vert \nabla L(\theta_1) + \theta_1 - \nabla L(\theta_2) - \theta_2\bigg\Vert^2\\
    =& \Vert  \nabla L(\theta_1) - \nabla L(\theta_2)\Vert^2 + \Vert \theta_1 - \theta_2 \Vert^2 + 2\langle \nabla L(\theta_1) - \nabla L(\theta_2) , \theta_1 - \theta_2   \rangle\\
    \leq & M_{L}^2 \Vert \theta_1 - \theta_2 \Vert^2 + \Vert \theta_1 - \theta_2 \Vert^2 - m \Vert \theta_1 - \theta_2 \Vert^2\\
    =& (1 - 2m + M_{L}^2) \Vert \theta_1 - \theta_2 \Vert^2
\end{align*}

Now we have ,
\begin{align*}
    & B < \frac{\sqrt{2m}}{L_{v}}\\
    &U_r < \frac{\sqrt{2m} - L_{v}B}{\bigg( 1 + \frac{2b}{1-k} \bigg)(L_{\mu}B + BL_{\pi} + K)}
\end{align*}

Hence , 
\begin{align*}
    &U_{r}\bigg( 1 + \frac{2b}{1-k} \bigg)(L_{\mu}B + BL_{\pi} + K) + L_{v}B < \sqrt{2m}\\
    \Rightarrow & M_{L} < \sqrt{2m}\\
    \Rightarrow & M_{L}^2 < 2m\\
    \Rightarrow & 1 - 2m + M_{L}^2 < 1
\end{align*}

\subsubsection{Proof of lemma \ref{martingale_ineq}}

We have 
\begin{align*}
     p_{m} &=  P \bigg(\bigg\{\sqrt{2 \bigg\vert\sum_{k = n_0}^{k = m}\xi(m,k+1) \beta_{k}(1 - \beta_{k}) \langle \theta_k - z_k , M_{k+1} \rangle \bigg\vert } >  \delta \bigg\}\bigg) \\
    &\qquad + P \bigg( \bigg\{ \sqrt{2 \bigg\vert \sum_{k = n_0}^{k = m}\xi(m,k+1) \beta_{k}(1 - \beta_{k}) \tilde{U}_{k+1}}\bigg\vert > \delta\bigg\}\bigg)\\
    &\qquad  + P \bigg( \bigg\{ \sqrt{\bigg(\frac{B}{d_2} + \frac{B}{d_1 \lambda}\bigg)\bigg\vert\sum_{k = n_0}^{k = m}\xi(m,k+1)\beta_k y_{k}  \tilde{Z}_{k+1}\bigg\vert} > \delta\bigg\}\bigg) \\
    &\qquad + P \bigg( \bigg\{\sqrt{\frac{4B}{d_1\lambda}\bigg\vert\sum_{k = n_0}^{k = m}\xi(m,k+1)\beta_k \langle z_k , \tilde{W}_{k+1}\rangle \bigg\vert } >  \delta \bigg\}\bigg)
\end{align*}

Now ,
\begin{align*}
     &(1 - \beta_{k}) \langle \theta_k - z_k , M_{k+1} \rangle\\
     \leq & 2 B (U_{\theta} + U_{v})U_{V}
\end{align*}

 We let $C = \exp(2 B (U_{\theta} + U_{v})U_{V} ), \varepsilon = 1, \zeta_{k,m} = \xi(m ,k+1)\beta_k, \gamma_1 = 1.$ Also,

 \begin{align*}
   \xi(m, k+1) &=   \prod_{i = k+1}^{m}(1 - \beta_i) \leq \prod_{i = k+1}^{m} e^{-\beta_i} = exp\bigg(-\sum_{k+1}^{m}\beta_i\bigg)\\
   &\qquad = exp\bigg(-\sum_{k+1}^{m}\frac{c_\beta}{(1+i)^{\sigma}}\bigg) \leq exp\bigg(-\sum_{k+1}^{m}\frac{c_\beta}{(1+i)}\bigg)\\
   &\qquad \leq exp \bigg( - \int_{k+1}^{m+1} \frac{c_\beta}{(y+1)}\,dy\bigg) = exp\bigg( c_\beta(\log(k+2) - \log(m+2))\bigg)\\
   &\qquad = exp\bigg( c_\beta \log \frac{k+2}{m + 2}\bigg) = \bigg( \frac{k+2}{m + 2}\bigg)^{c_\beta}
\end{align*}

Hence we have,
\begin{align*}
    &\max_{n_0 \leq k \leq m} \xi(m, k+1)\beta_k \leq \max_{n_0 \leq k \leq m} \bigg( \frac{k+2}{m + 2}\bigg)^{c_\beta} \frac{c_\beta}{(1+k)^{\sigma}} \leq \max_{n_0 \leq k \leq m} \bigg( \frac{3k}{m }\bigg)^{c_\beta} \frac{c_\beta}{(1+k)^{\sigma}}\\
    &\qquad \leq \max_{n_0 \leq k \leq m} \bigg( \frac{3k}{m }\bigg)^{c_\beta} \frac{c_\beta}{(k)^{\sigma}}
\end{align*}
 Now if $c_\beta \leq \sigma $, we have,
 \begin{align*}
     \max_{n_0 \leq k \leq m} \xi(m, k+1)\beta_k \leq 3^{c_\beta}c_\beta (n_0)^{c_\beta - \sigma}m^{-c_\beta}
 \end{align*}
Otherwise we have,
\begin{align*}
    \max_{n_0 \leq k \leq m} \xi(m, k+1)\beta_k  \leq 3^{c_\beta}c_\beta m ^{-\sigma}
\end{align*}

Let $\gamma_2 = 3^{c_\beta}c_\beta$. Also,
\[
\omega(m)=
\begin{cases}
(n_0)^{c_\beta - \sigma}m^{-c_\beta}, & \text{if } c_\beta \leq \sigma, \\
 m ^{-\sigma},  & \text{if } otherwise.
\end{cases}
\]

So we have,
\begin{align*}
     \max_{n_0 \leq k \leq m} \zeta_{k,m} \leq \gamma_2 \omega(m)
\end{align*}

So if we have, $\gamma_1  = 1, \epsilon = 1, C = \exp(2 B (U_{\theta} + U_{v})U_{V} ) $ and if we apply the result of theorem 2 of \cite{chandak2026concentrationboundtd0function}, then ,
\begin{align*}
    P\bigg(2 \bigg\vert\sum_{k = n_0}^{k = m}\xi(m,k+1) \beta_{k}(1 - \beta_{k}) \langle \theta_k - z_k , M_{k+1} \rangle \bigg\vert > \delta^2 \bigg) &\leq 2 e^{-\frac{D_1\delta^4}{\omega(m)}} , if \delta^2 \in \bigg(0, 4C\gamma_1/\epsilon \bigg]\\
    & 2 e^{-\frac{D_1\delta^2}{\omega(m)}} otherwise.
\end{align*}

Under the assumption that $\delta \leq 1$ we have,

\begin{align*}
     P\bigg(2 \bigg\vert\sum_{k = n_0}^{k = m}\xi(m,k+1) \beta_{k}(1 - \beta_{k}) \langle \theta_k - z_k , M_{k+1} \rangle \bigg\vert > \delta^2 \bigg) \leq 2 e^{-\frac{D_1\delta^4}{\omega(m)}}
\end{align*}
for some constant $D_1 > 0.$

Similarly under the assumption that $\delta \leq 1$ we have ,

\begin{align*}
    P \bigg( \bigg\{ \sqrt{2 \bigg\vert \sum_{k = n_0}^{k = m}\xi(m,k+1) \beta_{k}(1 - \beta_{k}) \tilde{U}_{k+1}}\bigg\vert > \delta\bigg\}\bigg) \leq 2 e^{-\frac{D_2\delta^4}{\omega(m)}}
\end{align*}
for some constant $D_2 > 0$,

\begin{align*}
    P \bigg( \bigg\{ \sqrt{\bigg(\frac{B}{d_2} + \frac{B}{d_1 \lambda}\bigg)\bigg\vert\sum_{k = n_0}^{k = m}\xi(m,k+1)\beta_k y_{k}  \tilde{Z}_{k+1}\bigg\vert} > \delta\bigg\}\bigg)\leq 2 e^{-\frac{D_3\delta^4}{\omega(m)}}
\end{align*}
for some constant $D_3 > 0$,

\begin{align*}
     P \bigg( \bigg\{\sqrt{\frac{4B}{d_1\lambda}\bigg\vert\sum_{k = n_0}^{k = m}\xi(m,k+1)\beta_k \langle z_k , \tilde{W}_{k+1}\rangle \bigg\vert } >  \delta \bigg\}\bigg) \leq 2 e^{-\frac{D_4\delta^4}{\omega(m)}}
\end{align*}
for some constant $D_4 > 0$.

Hence we have ,
\begin{align*}
    p_{n-1} \leq & 2 e^{-\frac{D_1\delta^4}{\omega(n-1)}} +2 e^{-\frac{D_2\delta^4}{\omega(n-1)}} +  2 e^{-\frac{D_3\delta^4}{\omega(n-1)}} + 2 e^{-\frac{D_4\delta^4}{\omega(n-1)}}\\
    \leq & 8e^{-\frac{D\delta^4}{\omega(n-1)}}
\end{align*}

where $D = \min\{ D_1, D_2, D_3, D_4\}$.

\subsubsection{Proof of lemma \ref{theta_bound}}

We have,

\begin{align*}
    &\Vert \theta_{t+1} - \theta^{*}\Vert^2\\
    =& \Vert \bar{\Gamma}(\theta_{t} + \beta_t \delta_t \nabla \log \pi_{\theta_t}(a_t|s_t)) - \theta^{*}\Vert^2\\
    \leq & \Vert \theta_{t} + \beta_t \delta_t \nabla \log \pi_{\theta_t}(a_t|s_t) - \theta^{*}\Vert^2\\
    =& \Vert  \theta_t - \theta^{*}\Vert^2 + 2 \beta_t \langle \theta_t - \theta^{*}, \delta_t \nabla \log \pi_{\theta_t}(a_t|s_t)  \rangle + \beta_t^2 \Vert  \delta_t \nabla \log \pi_{\theta_t}(a_t|s_t)  \Vert^2\\
    =& \Vert  \theta_t - \theta^{*}\Vert^2 + 2 \beta_t \langle \theta_t - \theta^{*}, (r_t - L_t + \phi(s_{t+1})^{\top} v_{t} - \phi(s_t)^{\top} v_{t}) \nabla \log \pi_{\theta_t}(a_t|s_t)  \rangle \\
    &\qquad + \beta_t^2 \Vert  \delta_t \nabla \log \pi_{\theta_t}(a_t|s_t)  \Vert^2\\
    =&  \Vert  \theta_t - \theta^{*}\Vert^2 + 2 \beta_t \langle \theta_t - \theta^{*}, (r_t - L(\theta_t) + V^{\theta_t}(s_{t+1}) - V^{\theta_t}(s_{t})) \nabla \log \pi_{\theta_t}(a_t|s_t)  \rangle \\
    &\qquad + \beta_t^2 \Vert  \delta_t \nabla \log \pi_{\theta_t}(a_t|s_t)  \Vert^2\\
    &\qquad + 2 \beta_t \langle \theta_t - \theta^{*}, (L(\theta_t) - L_t + \phi(s_{t+1})^{\top} v_{t} - V^{\theta_t}(s_{t+1}) - \phi(s_{t})^{\top} v_{t} + V^{\theta_t}(s_{t}))\nabla \log \pi_{\theta_t}(a_t|s_t)\rangle \\
    =&  \Vert  \theta_t - \theta^{*}\Vert^2 + 2 \beta_t \langle \theta_t - \theta^{*}, (r_t - L(\theta_t) + V^{\theta_t}(s_{t+1}) - V^{\theta_t}(s_{t})) \nabla \log \pi_{\theta_t}(a_t|s_t)  \rangle \\
    &\qquad + \beta_t^2 \Vert  \delta_t \nabla \log \pi_{\theta_t}(a_t|s_t)  \Vert^2\\
    &\qquad + 2 \beta_t \langle \theta_t - \theta^{*}, (L(\theta_t) - L_t + \phi(s_{t+1})^{\top} ( v_{t} - v^{*}(\theta_t)) - \phi(s_{t})^{\top}( v_{t} - v^{*}(\theta_t)) )\nabla \log \pi_{\theta_t}(a_t|s_t)\rangle \\
    &\qquad + 2 \beta_t \langle \theta_t - \theta^{*},(\phi(s_{t+1})^{\top}v^{*}(\theta_t) -V^{\theta_t}(s_{t+1}) - \phi(s_{t})^{\top}v^{*}(\theta_t) + V^{\theta_t}(s_{t}))\nabla \log \pi_{\theta_t}(a_t|s_t)  \rangle\\
     =&  \Vert  \theta_t - \theta^{*}\Vert^2 + 2 \beta_t \langle \theta_t - \theta^{*}, \EE_{\theta_t}[(r_t - L(\theta_t) + V^{\theta_t}(s_{t+1}) - V^{\theta_t}(s_{t})) \nabla \log \pi_{\theta_t}(a_t|s_t) ] \rangle \\
     &\qquad + 2 \beta_t \langle \theta_t - \theta^{*}, H(O_t, \theta_t) - \EE_{\theta_t}[H(O, \theta_t)]\rangle\\
    &\qquad + \beta_t^2 \Vert  \delta_t \nabla \log \pi_{\theta_t}(a_t|s_t)  \Vert^2\\
    &\qquad + 2 \beta_t \langle \theta_t - \theta^{*}, (L(\theta_t) - L_t + \phi(s_{t+1})^{\top} ( v_{t} - v^{*}(\theta_t)) - \phi(s_{t})^{\top}( v_{t} - v^{*}(\theta_t)) )\nabla \log \pi_{\theta_t}(a_t|s_t)\rangle \\
    &\qquad + 2 \beta_t \langle \theta_t - \theta^{*},(\phi(s_{t+1})^{\top}v^{*}(\theta_t) -V^{\theta_t}(s_{t+1}) - \phi(s_{t})^{\top}v^{*}(\theta_t) + V^{\theta_t}(s_{t}))\nabla \log \pi_{\theta_t}(a_t|s_t)  \rangle\\
      =&  \Vert  \theta_t - \theta^{*}\Vert^2 + 2 \beta_t \langle \theta_t - \theta^{*}, \nabla L(\theta_t) -\nabla L(\theta^{*})\rangle \\
     &\qquad + 2 \beta_t \langle \theta_t - \theta^{*}, H(O_t, \theta_t) - \EE_{\theta_t}[H(O, \theta_t)]\rangle\\
    &\qquad + \beta_t^2 \Vert  \delta_t \nabla \log \pi_{\theta_t}(a_t|s_t)  \Vert^2\\
    &\qquad + 2 \beta_t \langle \theta_t - \theta^{*}, (L(\theta_t) - L_t + \phi(s_{t+1})^{\top} ( v_{t} - v^{*}(\theta_t)) - \phi(s_{t})^{\top}( v_{t} - v^{*}(\theta_t)) )\nabla \log \pi_{\theta_t}(a_t|s_t)\rangle \\
    &\qquad + 2 \beta_t \langle \theta_t - \theta^{*},(\phi(s_{t+1})^{\top}v^{*}(\theta_t) -V^{\theta_t}(s_{t+1}) - \phi(s_{t})^{\top}v^{*}(\theta_t) + V^{\theta_t}(s_{t}))\nabla \log \pi_{\theta_t}(a_t|s_t)  \rangle\\
    \leq & \Vert  \theta_t - \theta^{*}\Vert^2  - 2m\beta_t \Vert \theta_t - \theta^{*}\Vert^2 +  2 \beta_t \langle \theta_t - \theta^{*}, H(O_t, \theta_t) - \EE_{\theta_t}[H(O, \theta_t)]\rangle + \beta_t^2 \Vert  \delta_t \nabla \log \pi_{\theta_t}(a_t|s_t)  \Vert^2\\
    &\qquad + 2\beta_t B\Vert \theta_t - \theta^{*}\Vert \vert L(\theta_t) - L_t \vert + 4\beta_t B \Vert \theta_t - \theta^{*}\Vert \Vert v_t - v^{*}(\theta_t)\Vert + 4\beta_t B \Vert \theta_t - \theta^{*}\Vert\epsilon_{\text{app}}\\
    \leq & (1 - 2 m \beta_t) \Vert \theta_t - \theta^{*}\Vert^2 +  2 \beta_t \langle \theta_t - \theta^{*}, H(O_t, \theta_t) - \EE_{\theta_t}[H(O, \theta_t)]\rangle + \beta_t^2 U_\delta^2 B^2\\
    &\qquad + \frac{2}{m}\beta_t B^2 U_{\delta}\vert L(\theta_t) - L_t \vert + \frac{4}{m}\beta_t B^2 U_{\delta}\Vert v_t - v^{*}(\theta_t) \Vert + \frac{4}{m}\beta_t B^2 U_{\delta}\epsilon_{\text{app}}
\end{align*}

Hence we have ,

\begin{align*}
    &\EE[\Vert \theta_{t+1} - \theta^{*}\Vert^2]\\
    \leq & (1 - 2 m \beta_t) \EE[\Vert \theta_t - \theta^{*}\Vert^2] +  2 \beta_t \EE\langle \theta_t - \theta^{*}, H(O_t, \theta_t) - \EE_{\theta_t}[H(O, \theta_t)]\rangle + \beta_t^2 U_\delta^2 B^2\\
    &\qquad + \frac{2}{m}\beta_t B^2 U_{\delta}\EE\vert L(\theta_t) - L_t \vert + \frac{4}{m}\beta_t B^2 U_{\delta}\EE\Vert v_t - v^{*}(\theta_t) \Vert + \frac{4}{m}\beta_t B^2 U_{\delta}\epsilon_{\text{app}}
\end{align*}

Now, analysing similar to lemma C.3 in \cite{wu2022finite}, we have,

\begin{align*}
    \vert \EE\langle \theta_t - \theta^{*}, H(O_t, \theta_t) - \EE_{\theta_t}[H(O, \theta_t)]\rangle \vert  = \mathcal{O}(1/t) + \mathcal{O}(\epsilon_{\text{app}}) 
\end{align*}

Hence we have ,

\begin{align*}
    &\EE[\Vert \theta_{t+1} - \theta^{*}\Vert^2]\\ 
    \leq & (1 - 2 m \beta_t) \EE[\Vert \theta_t - \theta^{*}\Vert^2] +  2 \beta_t (C_1/t + C_2 \epsilon_{\text{app}}) + \beta_t^2 U_\delta^2 B^2\\
    &\qquad + \frac{2}{m}\beta_t B^2 U_{\delta}\EE\vert L(\theta_t) - L_t \vert + \frac{4}{m}\beta_t B^2 U_{\delta}\EE\Vert v_t - v^{*}(\theta_t) \Vert + \frac{4}{m}\beta_t B^2 U_{\delta}\epsilon_{\text{app}}\\
    \Rightarrow & \EE[\Vert \theta_{t+1} - \theta^{*}\Vert^2]\\ 
    \leq & \underbrace{\xi(t,l) \EE\Vert \theta_l - \theta^{*}\Vert}_{I_1} + \underbrace{2\sum_{k = l}^{k = t}\Xi(t,k+1)\beta_k (C_1/k + C_2 \epsilon_{\text{app}}) }_{I_2}+ \underbrace{U_\delta^2 B^2\sum_{k = l}^{k = t}\Xi(t,k+1)\beta_k^2}_{I_3}\\
    &\qquad + \underbrace{\frac{2}{m} B^2 U_{\delta}\sum_{k = l}^{k = t}\Xi(t,k+1)\beta_k \EE\vert L(\theta_t) - L_t \vert}_{I_4} + \underbrace{\frac{4}{m} B^2 U_{\delta}\sum_{k = l}^{k = t}\Xi(t,k+1)\beta_k\EE\Vert v_t - v^{*}(\theta_t) \Vert}_{I_5}\\
    &\qquad + \underbrace{\frac{4}{m} B^2 U_{\delta}\epsilon_{\text{app}}\sum_{k = l}^{k = t}\Xi(t,k+1)\beta_k}_{I_6} 
\end{align*}

We have $ l = \min\{ i \geq 1 | 2m \beta_i < 1 \}$.

For term $I_1$ we have,
\begin{align*}
    &\Xi(t,l) \EE\Vert \theta_0 - \theta^{*}\Vert \\
    =& \prod_{k=l}^{k=t} (1-2m\beta_k)\EE\Vert \theta_0 - \theta^{*}\Vert\\
    \leq & \frac{U_\delta B}{m}\prod_{k=l}^{k=t} \exp(-2m \beta_k)\\
    =& \frac{U_\delta B}{m} \exp(-2m\sum_{k = l}^{k = t}\beta_k)\\
    =& \frac{U_\delta B}{m} \exp\bigg(-2mc_\beta \sum_{k = l}^{k = t}\frac{1}{1+k}\bigg) \\
    \leq & \frac{U_\delta B}{m} \exp\bigg(-2mc_\beta \int_{k = l}^{k = t + 1}\frac{1}{1+k}\bigg)\\
    =& \frac{U_\delta B}{m} \exp\bigg(2mc_\beta  (\log(l + 1)  -\log(t+2))  \bigg)\\
    =& \frac{U_\delta B}{m} \exp \bigg( 2mc_\beta \log \bigg(\frac{1+l}{t+2}  \bigg)\bigg)\\
    =& \frac{U_\delta B}{m} \bigg(  \frac{1+l}{t+2}\bigg)^{2mc_\beta}\\
    \leq & \frac{U_\delta B}{m} \bigg(  \frac{1+l}{t+2}\bigg)\\
\end{align*}
Now we have $2mc_\beta > l$. Hence $I_1 = \mathcal{O}(1/t).$

For term $I_2$ we have,
\begin{align*}
    I_2 = 2\sum_{k = l}^{k = t}\Xi(t,k+1)\beta_k (C_1/k + C_2 \epsilon_{\text{app}})
\end{align*}

Now, $\Xi(t,k+1) \leq \frac{k+2}{t}$. Hence we have,
\begin{align*}
    I_2 =& \frac{2 C_1}{t}\sum_{k = l}^{k = t}(k+2)\beta_k /k + \mathcal{O}(\epsilon_{\text{app}})\\
    =& \frac{2 C_1 c_\beta}{t}\sum_{k = l}^{k = t}(k+2) /(k (1+k))+ \mathcal{O}(\epsilon_{\text{app}})\\
    =& \mathcal{O}\bigg(  \frac{\log t}{t}\bigg) + \mathcal{O}(\epsilon_{\text{app}})\\
\end{align*}

For term $I_3$ we have,
\begin{align*}
    &U_\delta^2 B^2\sum_{k = l}^{k = t}\Xi(t,k+1)\beta_k^2\\
    \leq & \frac{U_\delta^2 B^2 c_\beta^2}{t}\sum_{k = l}^{k = t}(k+2)/(k+1)^2\\
    =& \mathcal{O}\bigg(  \frac{\log t}{t}\bigg)
\end{align*}

For term $I_4$ we have,
\begin{align*}
    I_4 &= \frac{2}{m} B^2 U_{\delta}\sum_{k = l}^{k = t}\Xi(t,k+1)\beta_k \EE\vert L(\theta_k) - L_k \vert \\
    &= \frac{2}{m} B^2 U_{\delta}\sum_{k = l}^{k = t}\sqrt{\Xi(t,k+1)^2\beta_k^2 (\EE\vert L(\theta_k) - L_k \vert)^2}\\
    &= \frac{2}{m} B^2 U_{\delta}\sum_{k = l}^{k = t}\sqrt{\Xi(t,k+1)\beta_k (\EE\vert L(\theta_k) - L_k \vert)^2}\sqrt{\Xi(t,k+1)\beta_k}\\
    &\leq \frac{2}{m} B^2 U_{\delta}\sqrt{\sum_{k = l}^{k = t}\Xi(t,k+1)\beta_k (\EE\vert L(\theta_k) - L_k \vert)^2}\sqrt{\sum_{k = l}^{k = t}\Xi(t,k+1)\beta_k}\\
    &\leq \frac{2c_\beta}{m\sqrt{m}} B^2 U_{\delta}\sqrt{\sum_{k = l}^{k = t}\frac{k+2}{t+2}\frac{1}{1+k} \EE\vert L(\theta_k) - L_k \vert^2}\\
     &\leq \frac{4c_\beta}{m\sqrt{m}} B^2 U_{\delta}\sqrt{\frac{1}{t}\sum_{k = l}^{k = t} \EE\vert L(\theta_k) - L_k \vert^2}
\end{align*}

Carrying out analysis similar to (\ref{ineq_I_3}), we have,
\begin{align}
    \sum_{k = l}^{k = t} \EE y_k^2 &\leq \underbrace{\sum_{k = l}^{k = t}\EE\bigg[\frac{y_k^2 - y_{k+1}^2}{2\gamma_k}\bigg]}_{I_a} +  \underbrace{\sum_{k = l}^{k = t} \EE[y_k(r_k-L(\theta_k))]}_{I_b} + \underbrace{\sum_{k = l}^{k = t}\EE\bigg[\frac{y_k}{\gamma_k}(L(\theta_k)-L(\theta_{k+1}))\bigg]}_{I_c}\notag\\
    &\qquad + \underbrace{\sum_{k = l}^{k = t}\EE\bigg[\frac{1}{\gamma_k}(L(\theta_k)-L(\theta_{k+1}))^2\bigg]}_{I_d} \notag\\ 
    &\qquad+  \underbrace{\sum_{k = l}^{k = t}\EE[\gamma_k(r_k-L_k)^2] }_{I_e}
\end{align}

Now, analysing terms $I_a - I_e$ similar to the analysis in section C.2 in \cite{wu2022finite}, we have,

\begin{align*}
    I_4 = \mathcal{O}(t^{-\nu/2}) + \mathcal{O}(t^{(\nu-1)/2}) + \mathcal{O}(\sqrt{\epsilon_{\text{app}}})
\end{align*}

Similarly, after analysing term $I_5$ similar to section C.3 in \cite{wu2022finite}, we have,
\begin{align*}
    I_5 = \mathcal{O}(t^{-\nu/2}) + \mathcal{O}(t^{(\nu-1)/2})  + \mathcal{O}(\sqrt{\epsilon_{\text{app}}})
\end{align*}

We take $\tau = 1$ for the results of  lemmas C.5 and C.7 in \cite{wu2022finite}.

For term $I_6$, we have,
\begin{align*}
   \frac{4}{m} B^2 U_{\delta}\epsilon_{\text{app}}\sum_{k = l}^{k = t}\Xi(t,k+1)\beta_k = \mathcal{O}(\epsilon_{\text{app}}) 
\end{align*}

Hence we have,
\begin{align*}
    &\EE[\Vert \theta_{t+1} - \theta^{*}\Vert^2] = \mathcal{O}\bigg(  \frac{\log t}{t}\bigg) + \mathcal{O}(t^{-\nu/2}) + \mathcal{O}(t^{(\nu-1)/2})  + \mathcal{O}(\epsilon_{\text{app}}) + \mathcal{O}(\sqrt{\epsilon_{\text{app}}})
\end{align*}

After simplification we have (ignoring the approximation error),
\begin{align*}
    &\EE[\Vert \theta_{t+1} - \theta^{*}\Vert^2] = \mathcal{O}(t^{-\nu/2}) + \mathcal{O}(t^{(\nu-1)/2})  
\end{align*}

\subsection{Hyperparameters}

Table \ref{tab:actor-critic-arch} gives us the information about the actor and critic network used in the experiments along with the learning rates. 

For CartPole, the feature map $\phi(s)$ consists of a constant bias term, the four normalized state variables, the squared values of these state variables, and two selected pairwise interaction terms. This results in a feature vector of dimension 11. The same feature map is used by both the actor and the critic. For FrozenLake, we use an exact one-hot representation of the state. Since there are 16 discrete states, each state is represented by a 16-dimensional feature vector with a value of one corresponding to the current state and zeros elsewhere. Consequently, the critic has one parameter for each state. For Blackjack, the feature map consists of a bias term, the normalized player sum, the normalized dealer card, an indicator for whether the player has a usable ace, the squared values of these state variables, and one selected interaction term. This gives a 7-dimensional feature vector, which is used by both the actor and the critic.

\begin{table}[t]
\centering
\caption{Actor and critic architecture and step-size (learning-rate) schedule used in each environment. }
\label{tab:actor-critic-arch}
\begin{tabular}{lccc}
\toprule
 & \textbf{CartPole-v1} & \textbf{FrozenLake-v1} & \textbf{Blackjack-v1} \\
\midrule
Actions $|A|$ & 2 (push left / right) & 4 (up/down/left/right) & 2 (stick / hit) \\ \hline
Actor parameterization & \makecell{binary logistic:\\ $\pi(a{=}0|s)=\sigma(\theta^\top\phi(s))$ \\ $\theta \in \mathbb{R}^{11}$} & \makecell{tabular softmax,\\ one logit anchored per state \\ $\theta \in \mathbb{R}^{3 \times 16}$} & \makecell{binary logistic \\ $\theta \in \mathbb{R}^{7}$} \\ \hline
Critic parameter $v$ & $v \in \mathbb{R}^{11}$ & $v \in \mathbb{R}^{16}$ & $v \in \mathbb{R}^{7}$ \\\hline
Feature map $\phi(s)$ & \makecell{bias, 4 normalized \\ state dims, their squares, \\ 2 cross terms} & \makecell{exact one-hot over \\ the 16 discrete states} & \makecell{bias, normalized player sum, \\ dealer card, ace indicator, \\ their squares, 1 cross term} \\ \hline
Approximation & linear FA &  tabular  & linear FA \\
\midrule
Critic step-size $\alpha_t$ & $5.0/(t+50)^{0.5}$ & $5.0/(t+50)^{0.5}$ & $1.0/(t+1)^{0.5}$ \\ \hline
Actor step-size $\beta_t$ & $5.0/(t+50)^{1.0}$ & $5.0/(t+50)^{1.0}$ & $2.0/(t+1)^{1.0}$ \\ \hline
Avg.-reward step-size $\gamma_t$ & $K\alpha_t,\ K{=}1.0$ & $K\alpha_t,\ K{=}1.0$ & $K\alpha_t,\ K{=}1.0$ \\ \hline
Critic projection radius $U_v$ & 50.0 & 50.0 & 50.0 \\
\bottomrule
\end{tabular}
\end{table}

\end{document}